\documentclass[lettersize,journal]{IEEEtran}
\usepackage{amsmath,amsfonts}
\usepackage{algorithmic}
\usepackage{algorithm}
\usepackage{array}
\usepackage[caption=false,font=normalsize,labelfont=sf,textfont=sf]{subfig}
\usepackage{textcomp}
\usepackage{stfloats}
\usepackage{url}
\usepackage{verbatim}
\usepackage{graphicx}
\usepackage{cite}
\usepackage{blindtext}
\usepackage{caption}
\usepackage{multirow}
\usepackage{bm}
\usepackage{booktabs}
\newcommand{\wwe}{$\mathcal{W}$-$\mathcal{W}^{+}$}
\newcommand{\Iskt}{\mathbf{I}_s}
\newcommand{\wplus}{\mathcal{W}^+}
\newcommand{\wspc}{\mathcal{W}}

\usepackage{float}
\usepackage{color}
\newcommand{\PAcc}{P\_Acc}

\begin{document}

\title{Semantics-Preserving Sketch Embedding for Face Generation}

\author{\author{Binxin Yang,~\IEEEmembership{Student Member,~IEEE},
Xuejin Chen, ~\IEEEmembership{Member,~IEEE}, 
Chaoqun Wang, Chi Zhang, Zihan Chen, ~\IEEEmembership{Student Member,~IEEE}, 
Xiaoyan Sun,~\IEEEmembership{Member,~IEEE}}}

\markboth{Journal of \LaTeX\ Class Files,~Vol.~14, No.~8, August~2021}%
{Shell \MakeLowercase{\textit{et al.}}: Semantics-Preserving Sketch Embedding for Face Generation}

\twocolumn[{
\renewcommand\twocolumn[1][]{#1}%
\maketitle
\begin{center}
    \centering
    \captionsetup{type=figure}
    \vspace{-0.5cm}
    \includegraphics[width=1\textwidth]{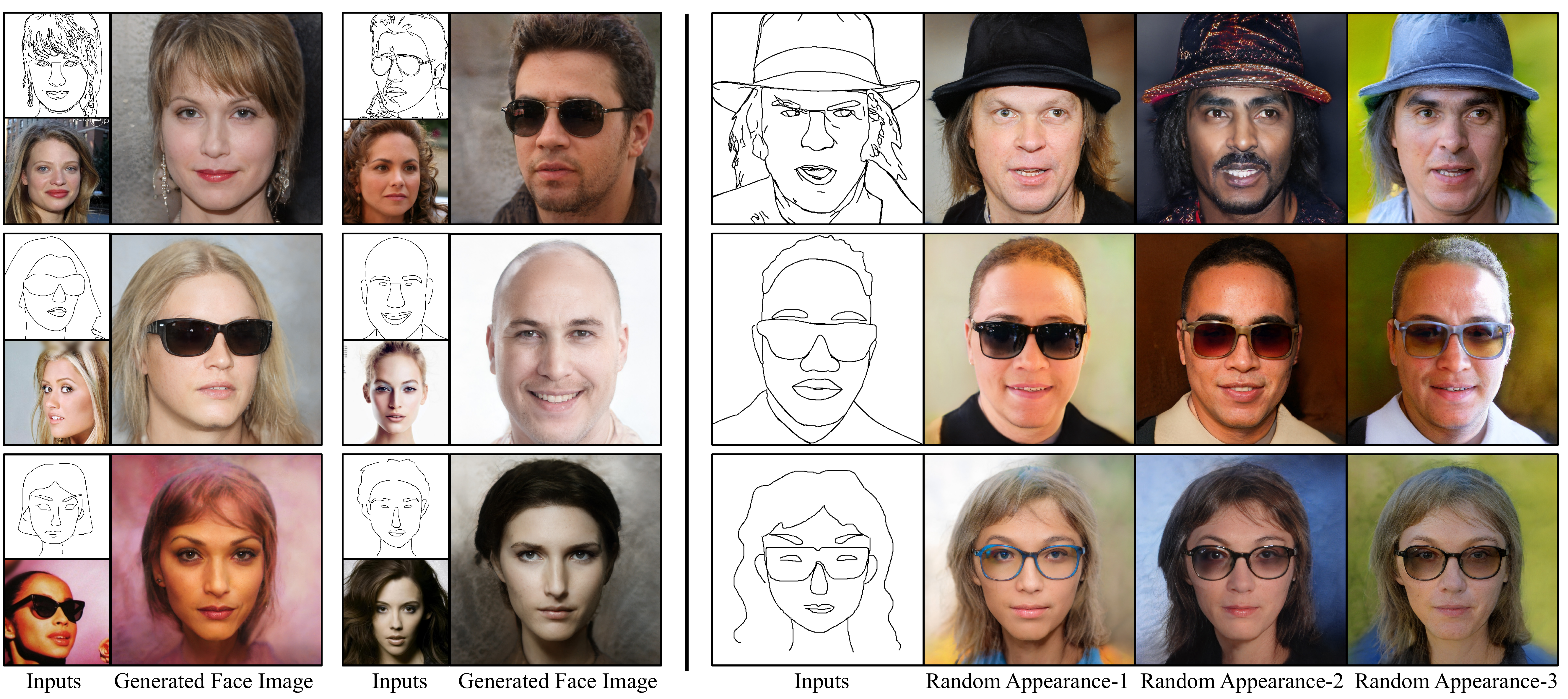}
    \captionof{figure}{Photorealistic face images generated from sketches in a large variety of styles with precise semantic control by our system. Our method effectively distills semantics from rough sketches and conveys it accurately to the latent space for a pre-trained face generator. From various sketches in edge style with fine details (first row), concise contour style (second row), and freehand style with geometric distortion (third row), our approach generates high-quality face images in many appearance modes while well-preserving the stroke semantics for all facial parts and accessories.}
\end{center}%
}]

{
  \renewcommand{\thefootnote}%
    {\fnsymbol{footnote}}
  \footnotetext[0]{This work was supported in part by the National Key R\&D Program of China under Grant No. 2020AAA0108600 and in part by the National Natural Science Foundation of China under Grants 62076230 and 62032006.}
  \footnotetext[0]{Xuejin Chen is the corresponding author.}
}

\begin{abstract}
With recent advances in image-to-image translation tasks, remarkable progress has been witnessed in generating face images from sketches.
However, existing methods frequently fail to generate images with details that are semantically and geometrically consistent with the input sketch, especially when various decoration strokes are drawn.
To address this issue, we introduce a novel $\mathcal{W}$-$\mathcal{W^+}$ encoder architecture to take advantage of the high expressive power of $\mathcal{W^+}$ space and semantic controllability of $\mathcal{W}$ space.
We introduce an explicit intermediate representation for sketch semantic embedding. With a semantic feature matching loss for effective semantic supervision, our sketch embedding precisely conveys the semantics in the input sketches to the synthesized images. 
Moreover, a novel sketch semantic interpretation approach is designed to automatically extract semantics from vectorized sketches.
We conduct extensive experiments on both synthesized sketches and hand-drawn sketches, and the results demonstrate the superiority of our method over existing approaches on both semantics-preserving and generalization ability.
\end{abstract}


\section{Introduction}
\IEEEPARstart{S}{ketch-based} face image generation aims to synthesize photo-realistic face images highly consistent with the input sketches, which can be widely applied to virtual human design, human-computer interaction, entertainment, and criminal investigation.
With the advent of image-to-image translation techniques~\cite{pix2pix,pix2pixHD,cyclegan}, tremendous progress has been made in sketch-based face image generation for various goals: the integrity of facial structure~\cite{Lines2facephoto}, robustness to freehand sketches~\cite{Deepfacepencil,DeepFaceDrawing}, and geometry-appearance disentanglement~\cite{Deepfaceediting}.

Inspired by the remarkable visual quality and fidelity of StyleGAN ~\cite{stylegan,stylegan2}, several recent approaches \cite{abdal2021styleflow,pixel2style2pixel,image2stylegan} invert an input into the learned latent $\mathcal{W}$ space of StyleGAN for synthesizing photo-realistic results.
In particular, the most representative method, pSp~\cite{pixel2style2pixel}, makes promising progress for sketch-based face image generation by encoding input sketches into $\mathcal{W^+}$ space of StyleGAN.
This high degree-of-freedom (DoF) $\mathcal{W^+}$ space is much more expressive for better reconstruction quality with less distortion. However, it also poses a challenge in precise control of desired attributes, especially for those of low distribution density when there are insufficient training examples, such as hats, glasses, etc.
Such a challenge leads to failures in reconstructing face images that are semantically consistent with sketch inputs, which have geometric distortions and ambiguous textures.
On the other hand, the original latent space $\mathcal{W}$ of StyleGAN, shows superiority in the semantic controllability and perceptual quality\cite{abdal2021styleflow,e4e}.

To map the semantics of input sketches into a correct region in the latent space without compromising reconstruction quality, we introduce a novel $\mathcal{W}$-$\mathcal{W^+}$ encoder architecture by combining the $\mathcal{W}$ space encoder with the $\mathcal{W^+}$ space encoder.
The $\mathcal{W}$ space encoder forces the encoded style codes to fall into the semantically meaningful manifold and stay close to the distribution of $\mathcal{W}$ for semantic consistency and perceptual quality.
Moreover, due to the natural sparseness of sketches, they are ambiguous in representing fine-level appearance. Directly embedding sketches in the high DoF $\mathcal{W^+}$ space leads to entanglement of high-level semantics and low-level details in the 18 style codes.
Previous studies \cite{stylegan2,pixel2style2pixel,abdal2021styleflow} have shown that the first 8-layer style codes control the geometry and semantics, which can be inferred from sketches.
Therefore, to ensure the precisely control of sketches over the semantics and geometry on generated images, we propose to only embed the first 8 layer style codes from input sketches and introduce a semantic feature matching (SFM) loss for effective supervision on semantics embedding.

Furthermore, to alleviate the difficulty for the encoders to capture the semantics from binary sketches that can not express the semantics explicitly, we propose a vector-based sketch semantic interpretation module to utilize the stroke structure and topological relationship between strokes for extracting semantics.
Such a module can capture the geometric and topological relationships common to edge maps and sketches, thus enabling better generalization capabilities.
Compared with previous approaches that train an end-to-end embedding network from raw sketches, our sketch-based face synthesis system utilizes an explicit intermediate representation of stroke semantics to support precise semantics control but does not require additional user inputs for specifying stroke categories.
In addition, our system is flexible enough to allow users to manually specify or modify the semantic category of single strokes to synthesize more diverse faces. 

In summary, the contribution of this paper includes:
\begin{itemize}
\item We propose a novel $\mathcal{W}$-$\mathcal{W^+}$ encoder architecture to take advantage of the semantic controllability and perceptual quality of $\mathcal{W}$ space and the reconstruction ability of $\mathcal{W^+}$ space, thus significantly improving the perceptual quality and semantical consistency with the input sketches.
\item We introduce the explicit semantic representation for sketch embedding. With this intermediate representation, our sketch embedding approach precisely conveys the semantics in the hand-drawn sketches for face synthesis and is well generalized to various styles of input sketches. 
\item We propose a GNN-based sketch semantic interpretation method that utilizes the stroke structure and topological relationship between strokes for better generalization ability to different types of sketches. 
 
\end{itemize}

\section{Related Work}
Generating images from sketches can be considered a specific image translation task. We will briefly introduce the development of image translation techniques and then sketch-based face image generation. Since we explicitly interpret the stroke semantics of the input sketch, we also discuss some related work of sketch semantic segmentation.

\subsection{Image to Image Translation}
The image-to-image translation task aims to translate images from one domain into another while preserving the structures and characteristics of inputs. 
Pix2pix~\cite{pix2pix} first explored conditional generative adversarial networks as a new paradigm for various image-to-image translation tasks like semantic segmentation, sketch-to-image generation, image colorization, etc.
Since then, many researchers followed Pix2pix and extended this new paradigm for many tasks.
One extension is to synthesize high-resolution images using a coarse-to-fine generator and a multi-scale discriminator~\cite{pix2pixHD}.
Without paired images in two domains for training, lots of unsupervised image-to-image translation networks have been proposed to adopt supervision at the domain level~\cite{unsupervisedI2I,cyclegan,DualGAN,DiscoGAN}.
Then many works have been studied to improve the diversity of image styles and scalability over multiple domains of the generative models.
However, these methods require careful network design as well as a long training period for one specific task. A generic image-to-image translation framework pSp~\cite{pixel2style2pixel} has been proposed to simplify the training process and demonstrates strong generalization ability by directly mapping the inputs into the $\mathcal{W^+}$ space.
To obtain better reconstruction results, Restyle~\cite{restyle} introduces a residual-based encoder to iteratively refine the latent code but with a negligible increase in inference time. On the other hand, instead of a time-consuming fine-tuning phase, HyperStyle~\cite{hyperstyle} uses a hypernetwork to predict offsets for modulating the generator weights.
Although existing networks have demonstrated extraordinary performance on image-to-image translation tasks, they still have limitations for some specific tasks, such as sketch-based face image generation.

\subsection{Sketch-based Face Image Generation}
Sketches can be considered as one specific type of 2D images. However, the sparsity and coarseness of hand-drawn sketches make significant differences from natural images that contain dense pixels and rich textures. With the great power of deep learning techniques, many efforts have been made to translate sketches into photos~\cite{sketchygan,sketchycoco,Ghosh_2019_ICCV,sketchhairsalon}.
However, the large diversity of the generated targets and the extreme sparsity of the sketches make generic image generation from sketches significantly challenging. 
Given a binary sketch image that describes the desired face structure, several approaches have been proposed to generate realistic face images\cite{DeepFaceDrawing,Deepfacepencil,Lines2facephoto,Deepfaceediting} and even videos~\cite{DeepFaceVideoEditing}. 
Besides binary sketches, extra information could be utilized for finer control on face generation, such as semantics masks in DrawingInStyles\cite{su2022drawinginstyles}, color palettes, light and shadow masks for controlling different attributes separately\cite{flexible}. However, the structure information of each stroke during the sketching process is lost when stored as a binary image.
Sketch understanding becomes more difficult due to the structural information loss, which is detrimental to the image generation based on sketches.
Therefore, we adopt the vector representation of sketches instead of a binary sketch image to better extract sketch features for stroke semantics interpretation.
Based on the explicit semantic representation, we train a new sketch encoder with a new training strategy to disentangle high-level semantics and fine-level appearances for face generation.

\subsection{Sketch Semantic Segmentation}
Semantic sketch segmentation aims to segment a sketch into groups with corresponding semantic labels.
Early approaches~\cite{HuangFL14,SchneiderT16} try to use traditional data-driven optimization algorithms, such as mixed integer programming and conditional random fields, for semantic segmentation.
With the rapid development of deep learning techniques, many studies have been conducted for deep sketch semantic segmentation.
Existing deep-learning-based methods can be divided into three categories: image-based~\cite{li2018fast,zhu20202d}, sequence-based~\cite{sketchgrouper,wu2018sketchsegnet,qi2019sketchsegnet+}, and graph-based~\cite{lumin2021sketchgnn}.
The image-based methods treat sketches as images and apply semantic image segmentation framework for sketch segmentation.
While these image-based methods ignore the stroke structural information, a post-processing step is usually added to avoid the points on a single stroke belonging to multiple category labels.
To utilize the structural information of hand-drawn sketches, sequence-based methods formulate sketch segmentation as a sequence prediction problem.
Although stroke structures are exploited, they fail to capture the spatial relationship when points are far from each other in the sequences.
To balance the structural and spatial context, SketchGNN~\cite{lumin2021sketchgnn} adopts a graph representation for sketches and formulates sketch segmentation as node classification.
It treats each point in a stroke as a node on the graph considering spatial information but ignores the sequential order of the points being drawn.
A post-processing step, such as graph-cut optimization, is also used to improve the consistency of semantic interpretations for the points on the same stroke. 
In contrast, our sequence-based stroke feature extraction and graph-based feature aggregation for sketch semantic segmentation exploit both the local stroke structure and global spatial correlations for semantic interpretation. 
\begin{figure}[ht]
\centering
\includegraphics[width=1\columnwidth]{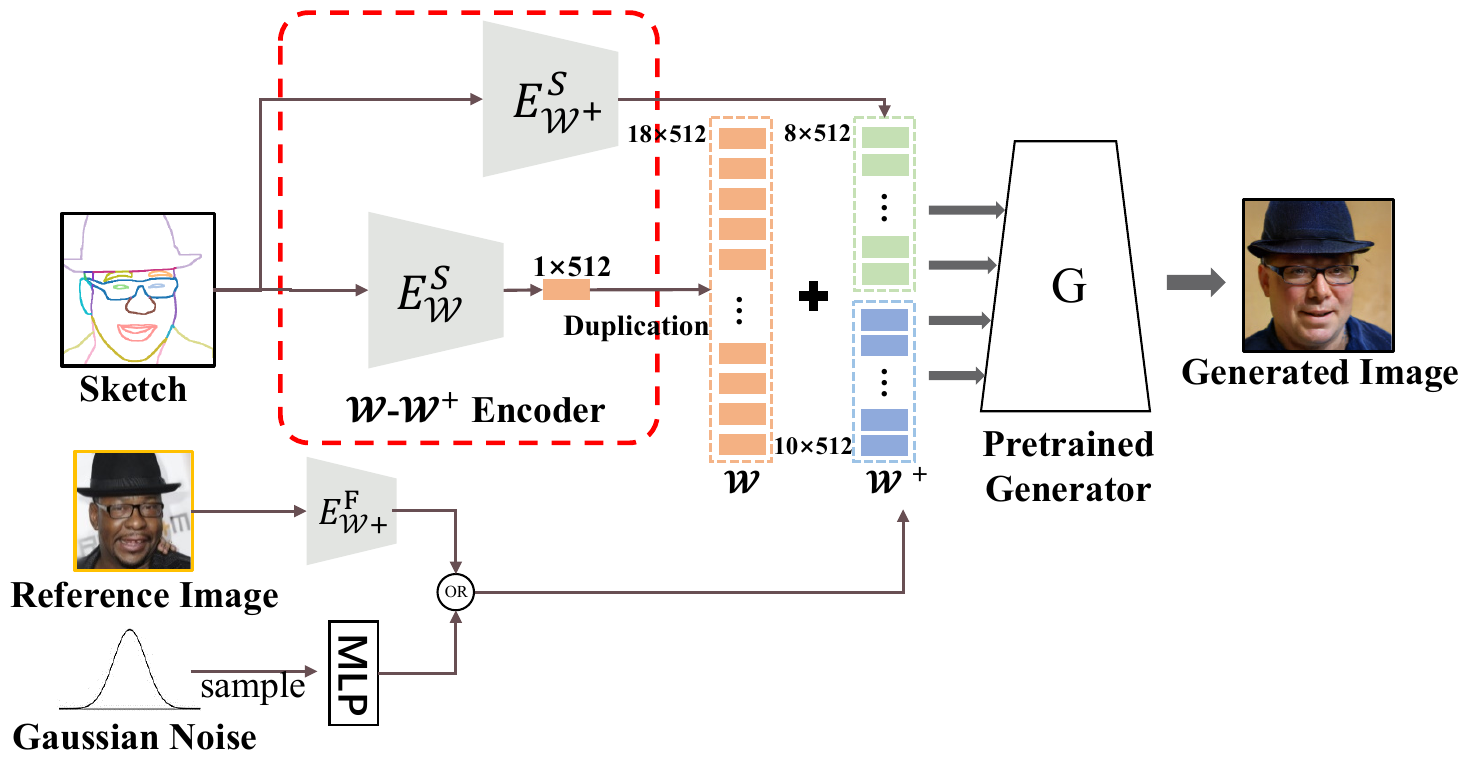}
\caption{Our semantics-preserving sketch embedding for face synthesis. 
To generate the 18 style codes for the pretrained StyleGAN generator $G$, our \wwe Encoder comprises two encoders that map the input sketch with stroke labels to the $\mathcal{W}$ and $\mathcal{W}^+$ latent spaces respectively. The low DoF $\wspc$ space ensures precise semantic controllability, while the high DoF $\wplus$ space ensures high perceptual quality and low distortion of the synthesized images. The fine-level appearance attributes can be controlled from a reference image and a randomly sampled latent vector. 
}
\label{fig:pipeline_1}
\vspace{-0.3cm}
\end{figure}

\section{Semantics-Preserving Sketch Embedding}
Our goal is to synthesize photo-realistic face images that are semantically consistent with input hand-drawn sketches.
Following the state-of-the-art techniques that use a learned latent space for face synthesis, we first invert the input sketch into a latent space and then synthesize a face image with a pre-trained StyleGAN generator. 
The pre-trained StyleGAN generator~\cite{stylegan} takes a latent code as input and maps it to 18 style codes for synthesis.
The 18 style codes separately represent high-level attributes (such as pose and face shapes) and fine-level stochastic variations (such as hairstyles, skin colors, etc). 
Many existing image manipulation approaches based on the pre-trained StyleGAN generator map the input into a latent space and then synthesize an image with the pretrained generator.
There are two options for the latent space of StyleGAN: $\mathcal{W}$ space and $\mathcal{W^+}$ space.
The characteristics of these two latent spaces have been comprehensively studied recently~\cite{zhu2020improved,e4e}.
The $\mathcal{W}^+$ space has better reconstruction capability but is semantically less controllable.
The $\mathcal{W}$ space leads to better perceptual quality and semantic controllability but shows more distortion compared with the original input. 

In our sketch-based image synthesis task, we want to take the advantages of semantic controllability and high perceptual quality from the $\mathcal{W}$ space while the high reconstruction capability from the $\mathcal{W+}$ space. 
Therefore, we propose a $\mathcal{W}$-$\mathcal{W}^+$ sketch embedding architecture, as Fig.~\ref{fig:pipeline_1} shows. 
Moreover, considering the fact that the sketches do not provide appearance information, mapping a sketch into the last 10 layers of $\mathcal{W}^+$ space is ambiguous, since those layers control the appearance of face images.
But sketches also convey fine geometric properties of the desired face such as eye shapes, expressions, hair boundaries, and even accessories like earrings. 
To solve the ambiguity of mapping sketches into fine appearance details while preserving geometric detail, we only embed the first eight style codes in the $\mathcal{W}^+$ space from the input sketch.

\subsection{Sketch2Face Synthesis Network Architecture}
Fig.~\ref{fig:pipeline_1} shows the architecture of our entire generation network with the proposed \wwe encoder for sketch semantic embedding. Our sketch-based face synthesis network consists of three encoders: a $\mathcal{W}$ sketch encoder $E_\mathcal{W}^S$, a $\mathcal{W^+}$ sketch encoder $E_\mathcal{W^+}^S$, and a $\mathcal{W^+}$ appearance encoder $E_\mathcal{W^+}^F$ to generate the fine-level appearance style codes optionally.
$E_\mathcal{W}^S$ encodes the input sketch $\Iskt$ as a $1\times512$ vector $\mathbf{w}_C=E_\mathcal{W}^S(\mathbf{I}_S)$ in the $\mathcal{W}$ space to present the coarse features of semantics and geometry.
The $E_{\mathcal{W}^+}^S$ encodes the fine-level geometric and semantic features of the sketches $\Iskt$ and outputs an $8\times512$ vector $\mathbf{w}_{FS}=E_{\mathcal{W}^+}^S(\mathbf{I}_S)$.
The missing 10 style codes for fine appearance attributes in $\wplus$ space can be compensated by a randomly sampled latent code or from a reference image. 
We also train a fine appearance encoder $E_\mathcal{W^+}^F$ to map a face image $\mathbf{I}_F$ to a $10\times512$ vector $\mathbf{w}_{FF}=E_\mathcal{W^+}^F(\mathbf{I}_F)$ of $\mathcal{W^+}$ space.
Without a reference image, the 10 appearance style codes can be randomly sampled from the $\mathcal{W}$ space.
The high-level style codes $\mathbf{w}_{FS}$ and the fine-level style codes $\mathbf{w}_{FF}$ are concatenated to form an $18\times512$ vector $\mathbf{w}_F$ in the $\wplus$ space.
To fuse $\mathbf{w}_C$ and $\mathbf{w}_F$, we duplicate $\mathbf{w}_C$ 18 times to form a $18\times512$ vector $\mathbf{w}^\prime_C$.
Then, the $\mathbf{w}^\prime_C$ and $\mathbf{w}_F$
are fed into a pretrained StyleGAN generator $G$ for synthesizing a high-quality face image $\mathbf{\hat{y}}=G(\mathbf{w})=G(\mathbf{\overline{w}}+\mathbf{w}^\prime_C+\mathbf{w}_F)$, where $\mathbf{\overline{w}}$ is the average style vector of pretrained generator.
In summary, this architecture balances the semantic controllability and high visual quality of $\mathcal{W}$ space with the expression ability of $\mathcal{W^+}$ space and enables more precise face generation by providing appearance features required for face from reference images.

\subsection{Loss Functions}
\label{method:OF}

We train our \wwe encoder in an end-to-end manner.
First, to provide supervision on texture details and perceptual similarities, we use $L_2$ loss and LPIPS~\cite{LPIPS} loss:
\begin{equation}
    L_2(\mathbf{x},\mathbf{\hat{x}})=||\mathbf{x}-\mathbf{\hat{x}}||_2,
\end{equation}
\begin{equation}
    L_{LPIPS}(\mathbf{x},\mathbf{\hat{x}})=||F(\mathbf{x})-F(\mathbf{\hat{x}})||_2,
\end{equation}
where $\mathbf{x}$ is the corresponding ground-truth face image of the input sketch, $\mathbf{\hat{x}}$ is the generated face image, and $F(\cdot)$ denotes the perceptual feature extractor. 
Following pSp~\cite{pixel2style2pixel}, to ensure the perceptual quality of the generated images, we also use a regularization loss, defined as:
\begin{equation}
    L_{reg}(\mathbf{w},\mathbf{\overline{w}})=||\mathbf{w}-\mathbf{\overline{w}}||_2,
\end{equation}
where $\mathbf{\overline{w}}$ is the average latent vector.
On the other hand, this regularization loss weakens the precise consistency between latent vector and the input sketch since it encourages the specific latent vector close to average vector of $\mathcal{W}$ space.
Considering that the perceptual quality is mainly influenced by the fine-level appearance, we assign larger weights to the last 10 codes that control the fine-level appearance and apply smaller weights to the first 8 codes that control geometry and semantics. 
This weighted regularization can enhance perceptual quality and avoid geometry inconsistency between the generated images and sketches.

Although the above losses can guide the network to generate high-quality face images, they can not guarantee the fidelity of high-level semantics of the input sketch, such as diverse facial parts and decorative accessories, in the synthesized image.
The primary reason is that both $L_2$ and $L_{LPIPS}$ focus primarily on fine texture details and perceptual similarities but do not provide effective supervision for high-level semantics.
We further introduce a semantic feature matching (SFM) loss to improve the semantic consistency between input sketches and synthesized face images.
Inspired by the perceptual loss~\cite{perceptual}, our semantic feature matching loss $L_{SFM}$ is designed to encourage smaller distance between the features of the generated image and the target image in semantic feature space.
We use a pretrained face segmentation network~\cite{CelebAMask-HQ} to extract high-level semantic features for both the generated faces and the ground-truth images.
For ease of presentation, we denote the $i$th-layer feature extractor of the pretrained face segmentation network as $F^{fs}_i$, which generates feature maps in size of $C_i\times H_i \times W_i$.
The semantic feature matching loss $L_{SFM}(\mathbf{x},\mathbf{\hat{x}})$ is then calculated as
\begin{equation} 
L_{SFM}(\mathbf{x},\mathbf{\hat{x}})=\sum_{i=1}^{T} \frac{1}{C_i\times H_i \times W_i} \|F^{fs}_i(\mathbf{x})-F^{fs}_i(\mathbf{\hat{x}})\|_2,
\end{equation} 
where $T$ is the total number of feature extraction layers.
The weighted combination of the above losses is used to train our sketch-based image generation network as:
\begin{equation}
\begin{split}
\label{eq:all-loss}
L_{S2F}(\mathbf{x},\mathbf{\hat{x}})=\lambda_1{L}_2(\mathbf{x},\mathbf{\hat{x}})+\lambda_2{L}_{LPIPS}(\mathbf{x},\mathbf{\hat{x}})+\lambda_3{L}_{SFM}(\mathbf{x},\mathbf{\hat{x}})\\+\lambda_4{L}_{Reg}(\mathbf{w}[:8],\overline{\mathbf{w}}[:8])+\lambda_5{L}_{Reg}(\mathbf{w}[9:],\overline{\mathbf{w}}[9:]).
\end{split}
\end{equation}

\section{Sketch Semantic Interpretation}
\label{method:SSI}

\begin{figure}[ht]
\centering
\includegraphics[width=1\columnwidth]{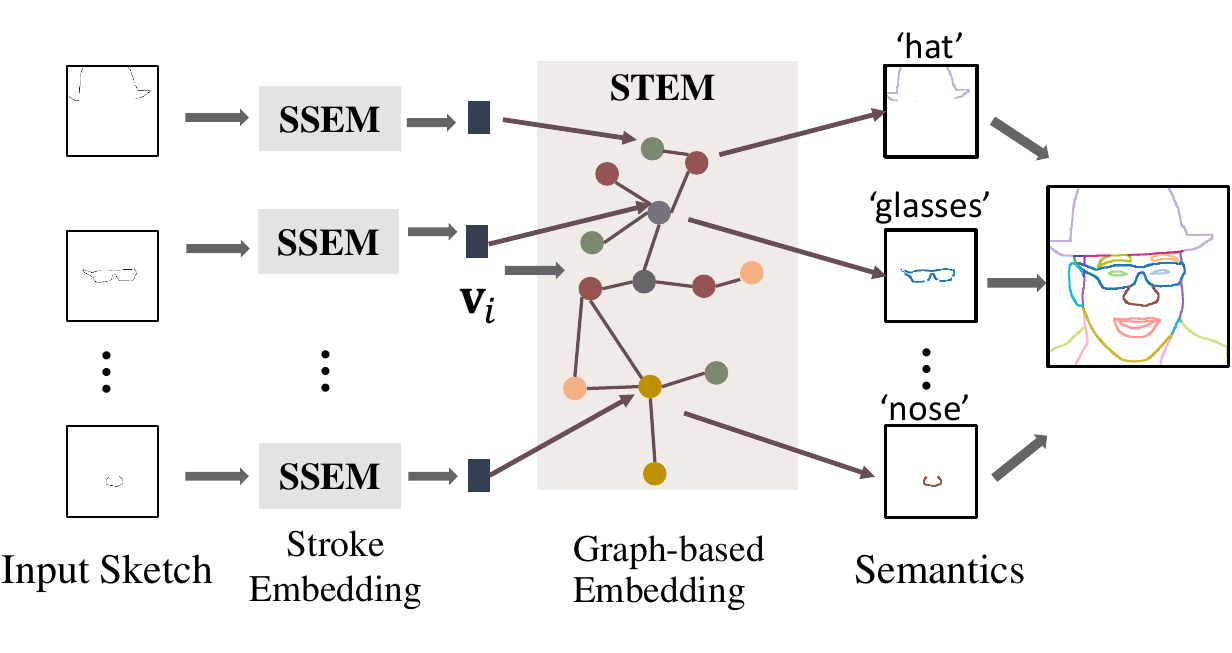}
\caption{Overview of our Sketch Semantic Interpretation Module. A vectorized sketch is first sent to the stroke-structure embedding module (SSEM) for extracting stroke structure features $\mathbf{v}_i$ from point sequences. Then, the sketch-topology embedding module (STEM) takes the stroke structure features $\mathbf{v}_i$ to build a graph and further aggregate them for extracting semantics.}
\label{fig:pipeline_2}
\vspace{-0.3cm}
\end{figure}

The goal of our \emph{Sketch Semantic Interpretation} (SSI) module is to assign each pixel on the sketch a semantic class label $c_{i}$ which represents its high-level semantics in the entire face.
During the drawing process, the sketches are stored as vectors, which contain the sequential order of points and which stroke each point belongs to.
Such information will facilitate the semantic interpretation of strokes. 
Therefore, we introduce a novel vector-based sketch semantic interpretation network, as shown in Fig.~\ref{fig:pipeline_2}.
The input of the sketch semantic interpreation module is a vectorial sketch $\mathcal{S}=\{S_i| i=1,2,...,N\}$ which consists of $N$ strokes $S_i$.
Each stroke $S_i$ consists of a sequence of points $P_i=\{(x_{ij},y_{ij})|j=1,\ldots,M_i\}$, where $M_i$ is the number of points on stroke $\mathbf{S}_i$ and $(x_{ij},y_{ij})$ are the 2D coordinates of $j$-th point of stroke $\mathbf{S}_i$. 
We first feed each stroke $\mathbf{S}_i$ into a stroke structure smbedding module (SSEM) to extract stroke-level structure features.
Then, the sketch topology embedding module (STEM) aggregates the stroke features with their spatial correlations based on a graph neural network to predict semantic labels for the strokes.

The SSEM consists of three layers of bidirectional-GRU~\cite{Bi-RNN,GRU}, 
which takes the point sequence $P_i$ of a stroke $S_i$ and its reverse-order counterpart $P_{i}^{r}$ as input and outputs a feature vector $\mathbf{v}_i\in \mathbb{R}^{K}$:
\begin{equation} \label{eq:stroke-feature}
\mathbf{v}_i=[{E_{\rightarrow}}(P_i);E_{\leftarrow}(P_{i}^{r})],
\end{equation}
where ${E_\rightarrow}$ and ${E_\leftarrow}$ are the GRU forward and backward encoders. 
$[\cdot;\cdot]$ denotes the concatenation operation.

To further utilize the topological relationship between strokes, we build a directed graph ${\mathcal{G}}=( \mathcal{V},\mathcal{E})$ for the input sketch. Each vertex $V_i \in \mathcal{V}$ represents a stroke $S_i$ carrying the stroke-level feature $\mathbf{v}_i$ as defined in Eq.~(\ref{eq:stroke-feature}).
For each vertex, we build $K$ directed edges from it to its $K$ nearest neighboring strokes.
Specifically, we calculate the centroid $\mathbf{q}_i$ of each stroke $S_i$ and look for its $K$ nearest neighbors according to the Euclidean distances between stroke centroids.
A directed edge $e(V_i\rightarrow V_j)$ will be constructed if the center $\mathbf{q}_j$ of a stroke $S_j$ is one of the $K$ nearest neighbors of $\mathbf{q}_i$.
A weight ${w}_{ij}$ is assigned to edge $e(V_i\rightarrow V_j)$ to describe their affinity as:
\begin{equation}
w_{ij}=MLP\big(\mathbf{q}_i,d(\mathbf{q}_i,\mathbf{q}_j),\theta(\mathbf{q}_i,\mathbf{q}_j)\big),
\end{equation}
where $d(\mathbf{q}_i,\mathbf{q}_j)$ is the distance between the two centroids $\mathbf{q}_i$ and $\mathbf{q}_j$ and $\theta(\mathbf{q}_i,\mathbf{q}_j)=\arctan(\frac{y_j-y_i}{x_j-x_i})$.

With the constructed graph $\mathcal{G}$ for the input sketch, we employ TAGConv~\cite{TAGconv} to aggregate stroke features taking advantage of its ability to aggregate long-range information on directed graphs.
This graph-based feature aggregation leads to more discriminative stroke features by considering the spatial correlations between neighboring strokes.   
At the end, a multi-layer perceptron is applied to the final layer feature $\mathbf{x}_i^L$ of vertex $V_i$ to predict its stroke label $\hat{c}_i$. 
Based on the predicted stroke-level semantic labels, we produce a semantic sketch by assigning each pixel with its predicted stroke label for our semantics-preserving sketch embedding network.

\section{Network Training}
\subsection{Dataset Preparation}
\label{sec:exp:dataset}

\begin{figure}[h]
\centering
\includegraphics[width=1\columnwidth]{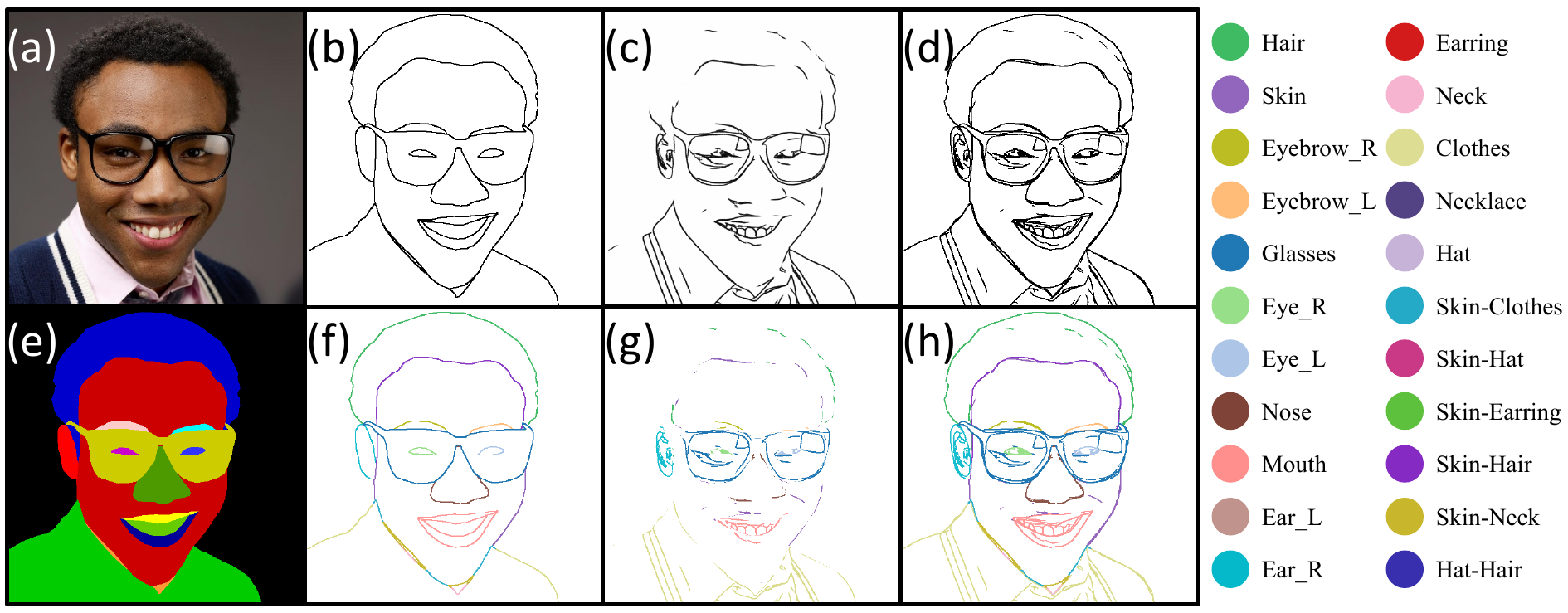}
\caption{Semantic sketch synthesis from face images. (a) A face image. (b) Extracted contour map from the semantic map. (c) Extracted edge map from the face image and (d) the fused sketch image by merging the edge and contour maps. (e) Its semantic map and (f) contour map with semantic labels. (g) Edge map with semantic labels and (h) the corresponding semantic sketch by merging (f) and (g). The 22 semantic labels are shown on the right. }
\label{fig:Sketch_synthesis}
\vspace{-0.2cm}
\end{figure}

We use CelebAMask-HQ~\cite{CelebAMask-HQ} as our raw dataset.
It contains 30,000 high-resolution face images and their segmentation maps of 19 semantic categories.  
We first follow pSp~\cite{pixel2style2pixel} to synthesize sketches from real face images and assign each pixel its semantic label according to the face segmentation map. 
Due to the rich textures in the face image, long strokes are usually missing in the synthesized edge maps, as Fig.~\ref{fig:Sketch_synthesis}(c) shows.
Moreover, the jagged edges around two adjacent semantic regions have inconsistent semantic labels. 
To produce cleaner strokes with consistent semantic labels, we further extract a contour map from the segmentation map.  
However, different from region-based segmentation maps, strokes typically indicate the borders of two adjacent regions.  
Therefore, we add six new semantic categories, including ‘skin-hair’, ‘skin-neck’, ‘skin-clothes’, ‘skin-hat’, ‘skin-earring’, and ‘hat-hair' to the boundary pixels. 
Since we focus on stroke semantics, we remove the ‘background' category and merge the ‘upper lip' and ‘lower lip' into the ‘mouth' category from the original 19 categories in the dataset.
Thus we get 22 stroke categories in total. 
When we merge the contour map and the edge maps, we dilate the contour map and filter out the noisy stroke labels on the edge map for overlapping pixels. The merged semantic sketch is then thinned to single-pixel width to mimic hand-drawn sketches. 

To obtain the final vectorized sketch data from a semantic sketch, we use a flood-fill algorithm~\cite{floodfill} to extract sequential points as strokes for each semantic category. 
While the centroid of a long and curvy stroke can not precisely represent its spatial information for stroke topology embedding, we cut each long stroke into short segments each of which contains 50 points at most.
An example vectorized sketch with semantic labels is shown in Fig.~\ref{fig:Sketch_synthesis} (h). 
Therefore, for each face image in the CelebAMask-HQ dataset, we have a quadruple of binary edge-sketch, semantic edge-sketch, binary contour-sketch, and semantic contour-sketch. 
The entire dataset is split into 90\% for training and 10\% for testing. 
In our experiments, we will evaluate our method on sketches of various styles. 
We denote the synthesized sketch dataset that consists of the contour-type sketches (Fig.~\ref{fig:Sketch_synthesis}(b) and (f)) as the CelebA-Contour dataset.
The dataset of the edge-type sketches (Fig.~\ref{fig:Sketch_synthesis}(d) and (h)) is denoted as CelebA-Edge.

\subsection{Training Details}

Since our semantics-preserving sketch embedding and sketch semantics interpretation are independent, we train the two modules separately. 
The networks are trained with the CelebA-Edge dataset only.
The whole framework is implemented with PyTorch and PyTorch Geometric.

\begin{figure*}[h]
\centering
\includegraphics[width=0.95\textwidth]{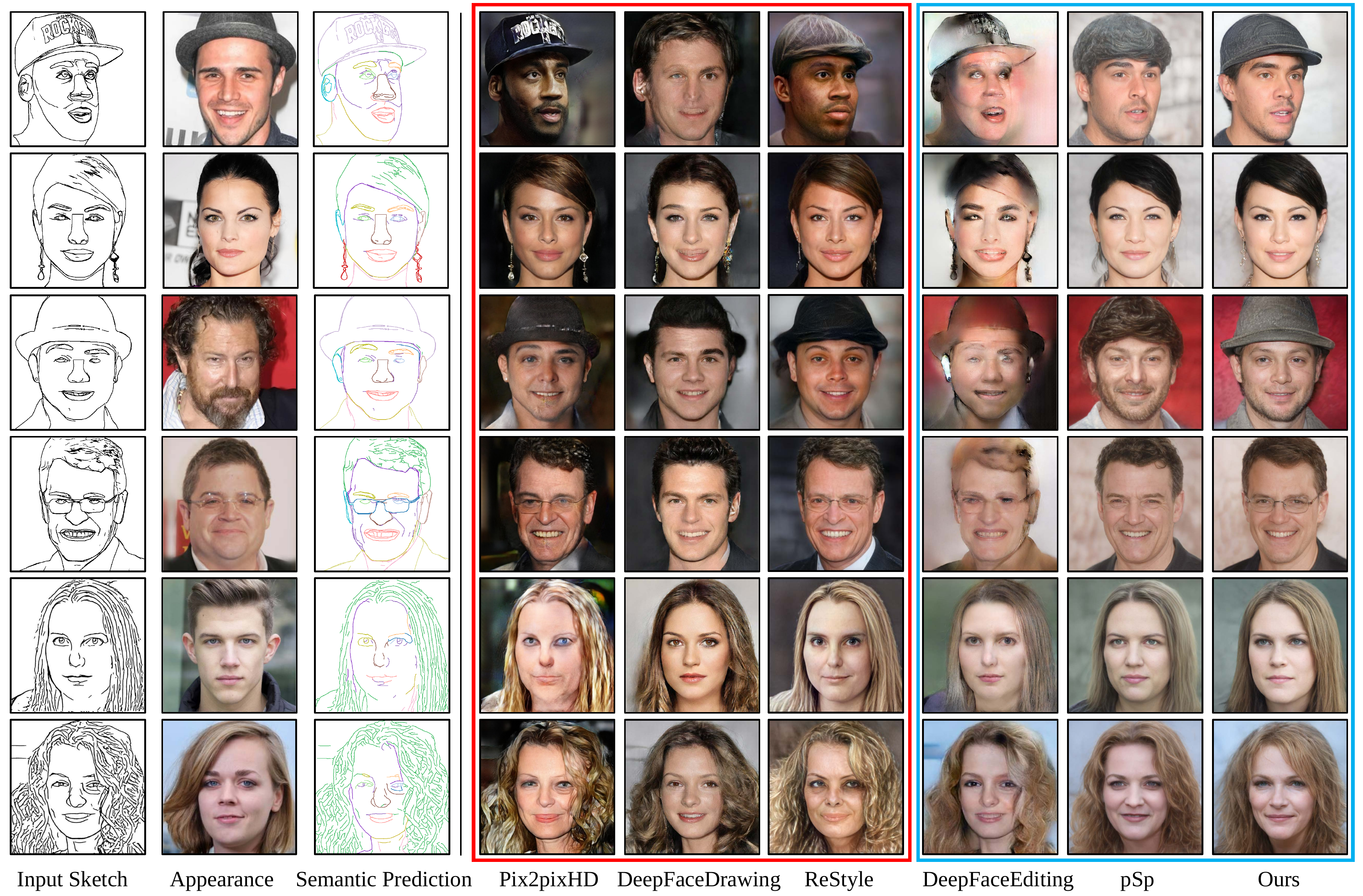}
\caption{Qualitative comparison of results from synthesized sketches. The sketches in the first four rows are from our CelebA-Edge dataset and the sketches in the last two rows are duplicated from DeepFaceEditing~\cite{Deepfaceediting}. While DeepFaceEditing requires a face image as the appearance reference, we also provide the same reference image to pSp and our method for style mixing (blue boxes). Appearance reference is not used in pix2pixHD, DeepFaceDrawing, and ReStyle (red
boxes). Our method is able to generate results that are semantically consistent with the input sketches (especially for hat, glasses, and earrings) in high perceptual quality.}
\label{fig:comparison_with_sota}
\vspace{-0.3cm}
\end{figure*}
\paragraph{Training of Semantic Sketch Embedding}
We use the synthesized semantic sketches and their corresponding face images to train our semantic sketch embedding network. 
There are much fewer images that contain accessories such as hats, glasses, earrings, and necklaces in the database. 
However, these semantic category minorities are important for synthesizing realistic portrait images. 
To alleviate the category imbalance, we augment the dataset by resampling images that contain hats, glasses, earrings, and necklaces by 23, 19, 3, and 16 times.
For the overall loss defined in Eq.~(\ref{eq:all-loss}), we set $\lambda_1=0.1$, $\lambda_2=0.8$, $\lambda_3=1$, $\lambda_4=0.00025$ and $\lambda_5=0.0025$.
We train the sketch embedding network using the Ranger~\cite{Ranger} optimizer with a constant learning rate of 0.001.
The training process takes about two days on one NVIDIA A100 GPU with 40GB GPU memory.

\paragraph{Training for Sketch Semantic Interpretation}
We train our sketch semantic interpretation network with the synthesized sketches described in Sec.~\ref{sec:exp:dataset}. 
Based on the graph representation of the entire sketch, we apply node classification on each vertex with the cross-entropy loss between predicted stroke labels and its synthesized semantic label. To improve the generalization ability on sketches of various levels of details, we use two simplified versions of sketches for data augmentation by retaining only the top 3 and the top 10 longest strokes for each semantic category in one sketch.
The augmented dataset contains 81k and 27k simplified sketches of the two levels respectively.
Adam~\cite{adam} optimizer is used with $\beta_1=0.9,\beta_2=0.999$, while the learning rate is set to 0.001 and decayed by $\gamma=0.98$ each epoch.
The batch size is 10.
The training process of the SSI module takes about 22 hours on an NVIDIA V100 GPU with 32GB GPU memory.

\section{Experiments}
In this section, we show extensive experimental results, comparing our method with state-of-the-art approaches both quantitatively and qualitatively and further analyzing the effect of each proposed component of our method.
\subsection{Comparison with Existing Methods}
\begin{figure}[h]
\centering
\includegraphics[width=1\columnwidth]{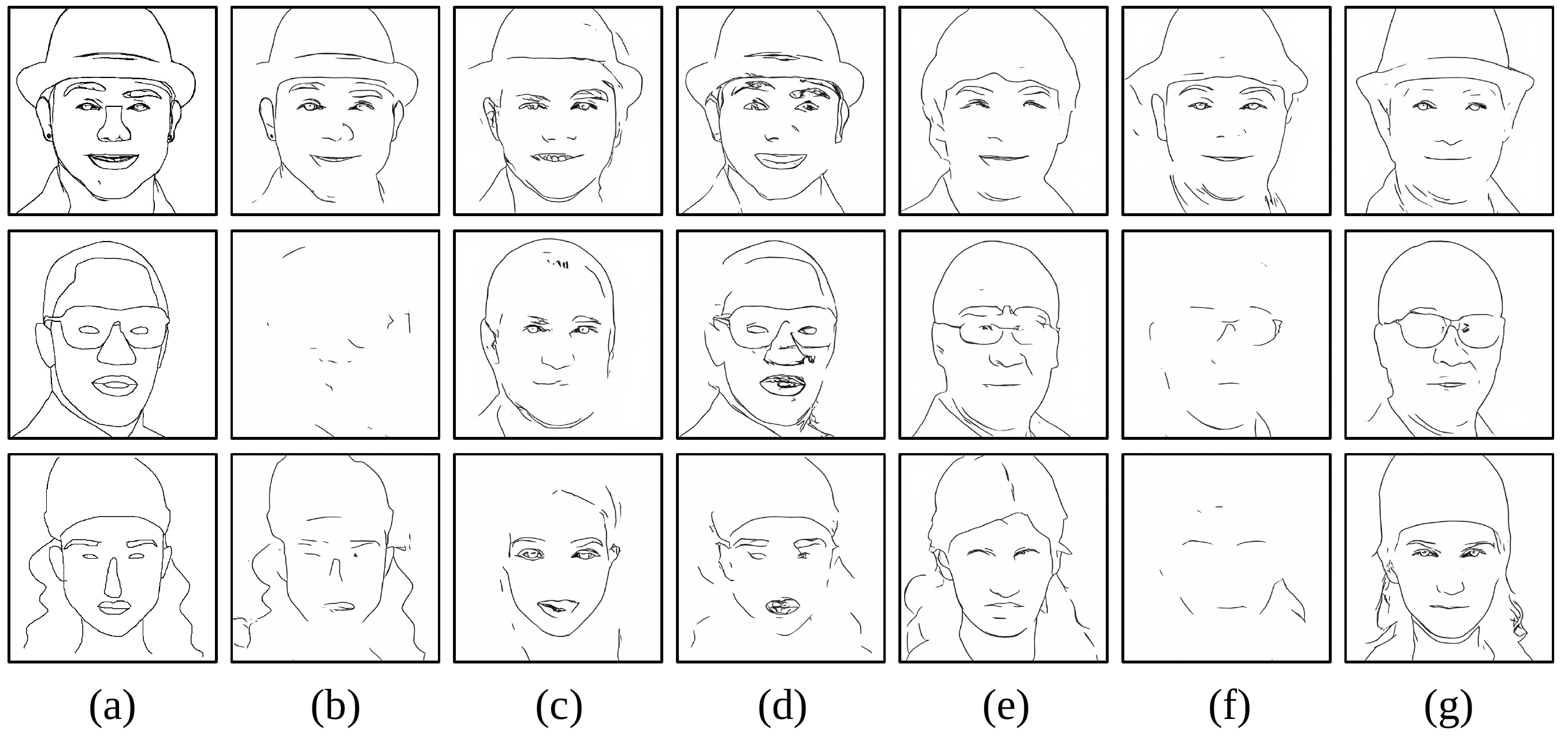}
\caption{Visual comparison of geometry preservation with other methods. (a) Input sketches. (b) Results of pix2pixHD. (c) Results of DeepFaceDrawing. (d) Results of DeepFaceEditing. (e) Results of pSp. (f) Results of ReStyle. (g) Our results. The synthesized sketches are from the generated images randomly selected in Fig.~\ref{fig:comparison_with_sota} and Fig.~\ref{fig:cp_freehand}. }
\label{fig:geometry_preservation}
\vspace{-0.3cm}
\end{figure}
We compare our method with five state-of-the-art sketch-based face image generation methods, 
namely pix2pixHD~\cite{pix2pixHD}, DeepFaceDrawing~\cite{DeepFaceDrawing}, DeepFaceEditing~\cite{Deepfaceediting}, pSp~\cite{pixel2style2pixel}, and ReStyle~\cite{restyle} for both quantitative and qualitative evaluation.
For a fair comparison, we re-train pix2pixHD, pSp, and ReStyle using the binary sketches and face images from the CelebA-Edge dataset. 
For DeepFaceDrawing and DeepFaceEditing, since no official code for network training and sketch synthesis is available, we use their pretrained models with default settings.

\begin{figure*}[h]
\centering
\includegraphics[width=0.95\textwidth]{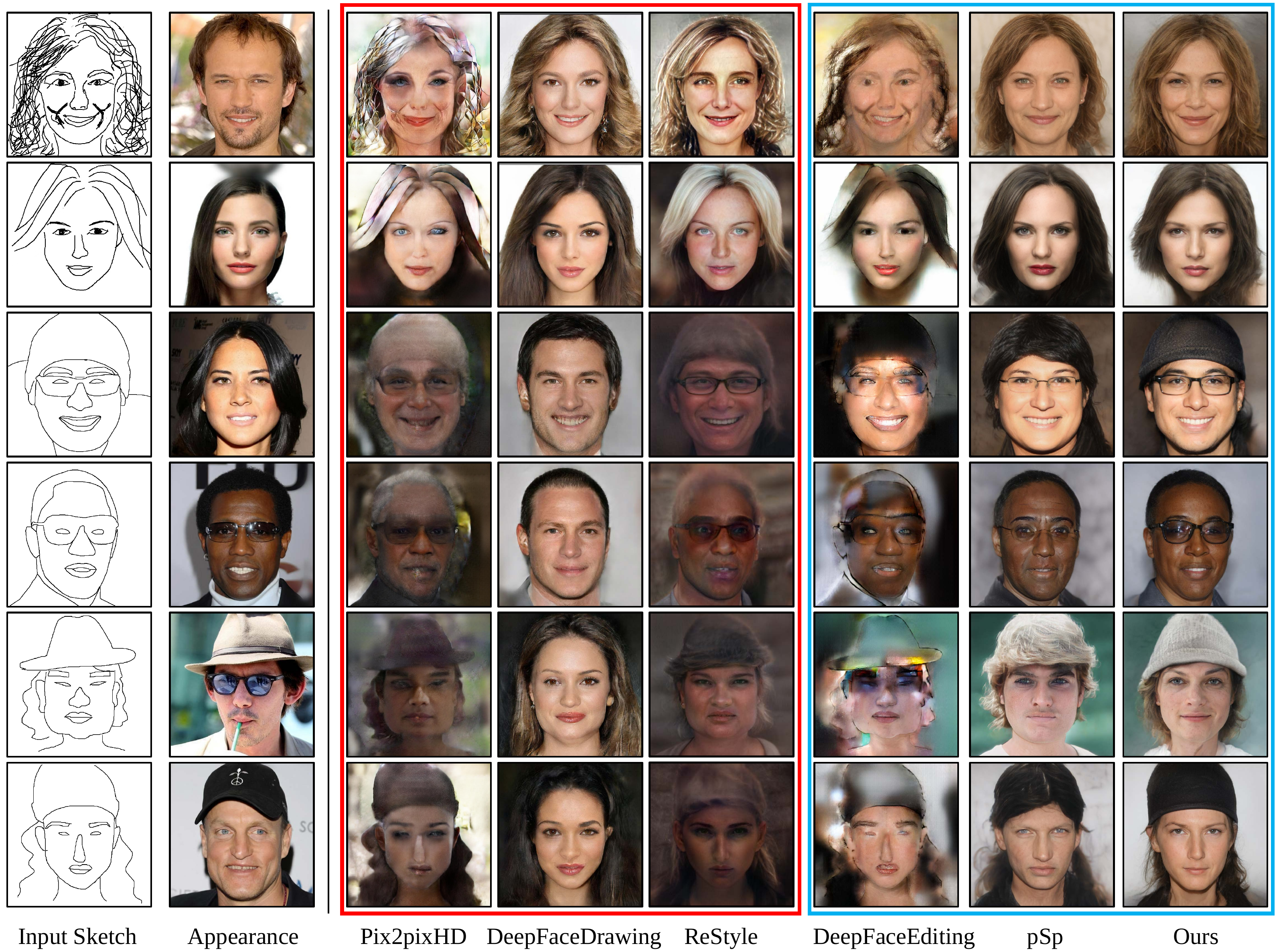}
\caption{Face generation from freehand sketches. The sketches in the top, middle and bottom two rows are from the paper of DeepFaceDrawing, our CelebA-Contour dataset, and our freehand dataset.
The red and blue dashed boxes indicate respectively the absence and presence of appearance reference images in the inputs.
Compared to pix2pixHD, DeepFaceDrawing, ReStyle, and DeepFaceEditing, our method shows better generalization to freehand sketches.
The results of our method and pSp are comparable in terms of realism.
But our results are more semantically consistent with the input sketches, especially for hats and hairstyles.}
\label{fig:cp_freehand}
\vspace{-0.4cm}
\end{figure*}

\paragraph{Quantitative Analysis}
We adopt the four metrics, including FID~\cite{FID} to evaluate the quality of synthesized images, LPIPS~\cite{LPIPS} to evaluate the similarity between the synthesized images and GT images, \PAcc~for semantic accuracy in the synthesized images, and Chamfer Distance (CD)~\cite{barrow1977parametric} for geometry preservation. 
We define \PAcc~as the pixel-wise accuracy between the predicted semantic map using the pretrained face segmentation network on the generated face image~\cite{CelebAMask-HQ}
and its corresponding ground-truth semantic map. The geometry preservation performance is evaluated by measuring the Chamfer Distance between the input sketch and the sketch synthesized from the generated face image, as mentioned in Sec.~\ref{sec:exp:dataset}.
Tables \ref{tab:quantitative_edge} and \ref{tab:quantitative_contour} present the quantitative evaluation results.
Because pix2pixHD~\cite{pix2pixHD} essentially transfers local textures based on pixel-wise correspondence, it achieves great results on FID, LPIPS, \PAcc, and CD on the synthesized sketches from the CelebA-Edge dataset, which has the same style as the training data.  
However, when tested on the contour-style sketches from the CelebA-Contour dataset which represents a different sketch style, its performance on all metrics drops significantly, showing its poor generalization ability on different types of sketches.
DeepFaceDrawing~\cite{DeepFaceDrawing} performs consistently on both edge and contour datasets because it maps various types of sketches into meaningful facial component manifolds by manifold projection. 
However, this manifold projection approach inevitably results in losing some of the local geometric information, thus DeepFaceDrawing struggles with geometry preservation.
Although DeepFaceEditing~\cite{Deepfaceediting} shows remarkable results on CD, it fails to produce high-quality images from sketches that are different from training data.
Compared with the latent-space-based method pSp~\cite{pixel2style2pixel} and ReStyle\cite{restyle}, our approach achieves comparable results on the CelebA-Edge dataset.
But on the CelebA-Contour dataset, our approach outperforms pSp and ReStyle in terms of P\_Acc, FID, LPIPS, and CD.
Particularly, the iterative refinement mechanism of ReStyle enlarges the gap between the edge-style and contour-style sketches, eventually leading to a significant performance degradation of ReStyle on CelebA-Contour. Our approach, on the other hand, performs consistently well on the CelebA-Contour, demonstrating that our representation of semantic sketches can effectively reduce the domain gap between various styles of sketches for high-quality face image generation. 

\begin{table}[h]
\centering
\caption{Quantitative comparison on CelebA-Edge.}
\begin{tabular}{@{}lcccc@{}}
\toprule
Method                        & P\_Acc ($\uparrow$)        & FID ($\downarrow$)  & LPIPS ($\downarrow$)  & CD ($\downarrow$)      \\ \midrule
pix2pixHD               & \textbf{0.942}            & \textbf{20.61}    & \textbf{0.403} & 0.00064      \\
DeepFaceDrawing         & 0.802    & 57.16    & 0.507   & 0.00219       \\
DeepFaceEditing         & 0.634    & 121.60   & 0.617   & \textbf{0.00016}        \\
pSp                     & 0.891    & 42.53    & 0.448   & 0.00161       \\
ReStyle                     & 0.913    & 34.75   & 0.523   & 0.00097     \\
Ours                    & 0.902    & 38.89    & 0.446   & 0.00166           \\ \bottomrule
\end{tabular}
\label{tab:quantitative_edge}
\end{table}

\begin{table}[h]
\centering
\caption{Quantitative comparison on CelebA-Contour.}
\begin{tabular}{@{}lcccc@{}}
\toprule
Method                        & P\_Acc ($\uparrow$)        & FID ($\downarrow$)  & LPIPS ($\downarrow$)  & CD ($\downarrow$)      \\ \midrule
pix2pixHD               & 0.850    & 87.89    &0.635 & 0.02886\\
DeepFaceDrawing         & 0.786    & 67.45    & 0.530  & 0.00502        \\
DeepFaceEditing         & 0.607    & 137.28   & 0.678  & \textbf{0.00050}         \\
pSp                     & 0.878    & 51.28    & 0.472     & 0.00385     \\
ReStyle            & 0.864    & 75.17    & 0.767    & 0.00375     \\
Ours                    & \textbf{0.892}    & \textbf{43.18}    & \textbf{0.466}     & 0.00311         \\ \bottomrule
\end{tabular}
\label{tab:quantitative_contour}
\vspace{-0.2cm}
\end{table}

\paragraph{Qualitative Analysis}
We further show qualitative comparisons of these methods in Fig.~\ref{fig:comparison_with_sota}.
For a fair comparison, besides the sketches from our dataset (the first four rows), we also duplicate the sketches presented in DeepFaceEditing (the last two rows) as input. 
As Fig.~\ref{fig:comparison_with_sota} shows, Pix2pixHD works quite well for the synthesized sketches with well-aligned geometry and fine appearance details, especially for the earrings.
This benefits from the rich details in the input sketches synthesized from real images and the pixel-to-pixel correspondence of the pix2pixHD model. 
But as for the sketches from DeepFaceEditing, pix2pixHD fails to generate realistic results due to its poor generalization performance. 
DeepFaceDrawing enhances its generalization ability for various sketches by encoding the input sketch into a latent space and performing manifold projection.
But such a projection inevitably loses details and leads to results that are not precisely consistent with the input sketches. Since the face manifold in DeepFaceDrawing is mainly built for frontal faces, it fails to synthesize faces of different poses (first row) or accessories (earrings in the second row and glasses in the fourth row).
DeepFaceEditing disentangles the geometry and appearance, which enables accurate control of the geometry and appearance by sketches and reference face images.
However, while designed primarily for local face editing, DeepFaceEditing is poorly generalized to different types of sketches (the first four rows).
The pSp and ReStyle models trained for sketch-based image synthesis can generate highly realistic images but frequently fail to correctly generate particular accessories, such as hats, earrings, and glasses. This is because of the tradeoff between geometric alignment and perceptual quality in the synthesized images. The high DoF of the $\wplus$ space supports its expressiveness for the densely distributed regions in the latent space. However, when the training samples are insufficient, for examples, the face images with hat, glasses, or earrings, inaccurate mapping occurs in such a high DoF space.
In contrast, our model enables more precise semantic control in face synthesizing while maintaining the high perceptual quality of the generated face images and generalization ability to various sketch inputs.
We also visually evaluate the performance of different methods on geometry preservation in Fig.~\ref{fig:geometry_preservation} by comparing the input sketches and the extracted sketches from synthesized images, which are randomly selected from Fig.~\ref{fig:comparison_with_sota} and Fig.~\ref{fig:cp_freehand}.
The pix2pixHD and DeepFaceEditing can generate images that precisely align with the input sketches in terms of geometry, but they fail to synthesize realistic images from sketches that are different from training data, as shown in Fig.~\ref{fig:comparison_with_sota}.
The generated image of pix2pixHD on the CelebA-Contour is precisely aligned with the sketch, but it is so blurry that most of the edges cannot be detected, as shown in the second row of Fig.~\ref{fig:geometry_preservation}.
The remaining methods, including ours, typically map sketches into a learned latent space for generalization ability to various types of sketches, thus inevitably losing some of the local geometric information. Actually, this situation is beneficial for non-expert users because sketches from them are usually of varying degrees of abstraction. If the results are exactly geometrically consistent with some abstract sketches, the results would be unrealistic.
In terms of geometry preservation, pSp and our method outperform DeepFaceDrawing among these latent-space-based methods.
This is because our latent space is more semantically meaningful than the one of DeepFaceDrawing, thus our latent space can express a greater variety of geometries.
The iterative refinement mechanism of ReStyle progressively enlarges gaps between the training and testing data in the latent space, which ultimately results in blurry outputs whose edges are hard to be detected. 
  
\paragraph{Evaluation for Various Sketch Styles}
The generalization capability to various types of sketches is highly demanded by sketch-based face image generation. 
To evaluate the generalization ability of each method, we conduct qualitative experiments on different sketches.
Fig.~\ref{fig:cp_freehand} show the synthesized results from various sketches, including two hand-drawn sketches presented in the DeepFaceDrawing paper (top two rows), two synthesized sketches from our CelebA-Contour dataset (mid two rows), and two freehand sketches (bottom two rows). 
These sketch inputs are significantly different from the sketches in CelebA-Edge used for training.
We construct a freehand sketch dataset by inviting thirty non-expert volunteers to draw freehand face sketches. Compared to the synthesized edge sketches that are geometrically accurate, there are various levels of distortions in freehand sketches drawn by common users. 
As shown in Fig.~\ref{fig:cp_freehand}, the results generated by pix2pixHD and DeepFaceEditing exhibit strong geometric correspondences between the sketches and face images but the perceptual quality is unsatisfying because of geometric distortions and deficiency of details. 
The manifold projection of DeepFaceDrawing ensures the image realism but limits the diversity. It fails to produce face images that are consistent with the input sketches of various poses, face shapes, hairstyles, and accessories.
The increasing gap caused by the iterative refinement mechanism of Restyle results in blurry outputs from sketches in different styles with the edge styles used for training.
Although pSp is capable of generating highly realistic face images, the results are sometimes not semantically consistent, such as missing hats and glasses, etc.
In comparison, our method can generate face images that are more semantically consistent with the input sketches while maintaining the realism of generated faces, especially for accessories.

\subsection{Ablation Study}
\begin{figure}[h]
\centering
\includegraphics[width=1\columnwidth]{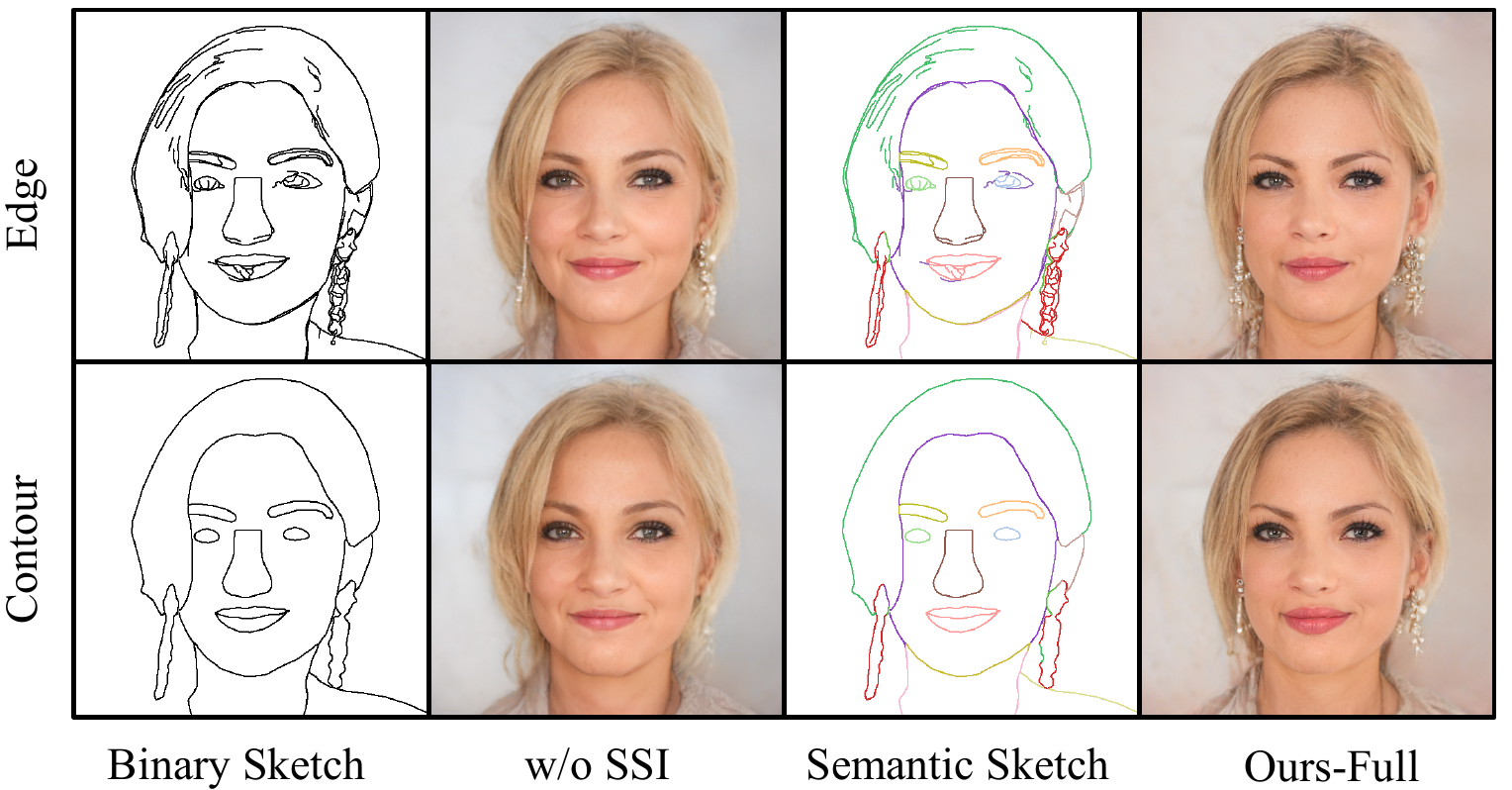}
\caption{Results of “w/o SSI” model variant and our full model. The SSI module improves the generalization ability in terms of sketches with different level of details.}
\label{fig:ablation-SSI}
\end{figure}

\begin{figure*}[h]
\centering
\includegraphics[width=0.95\textwidth]{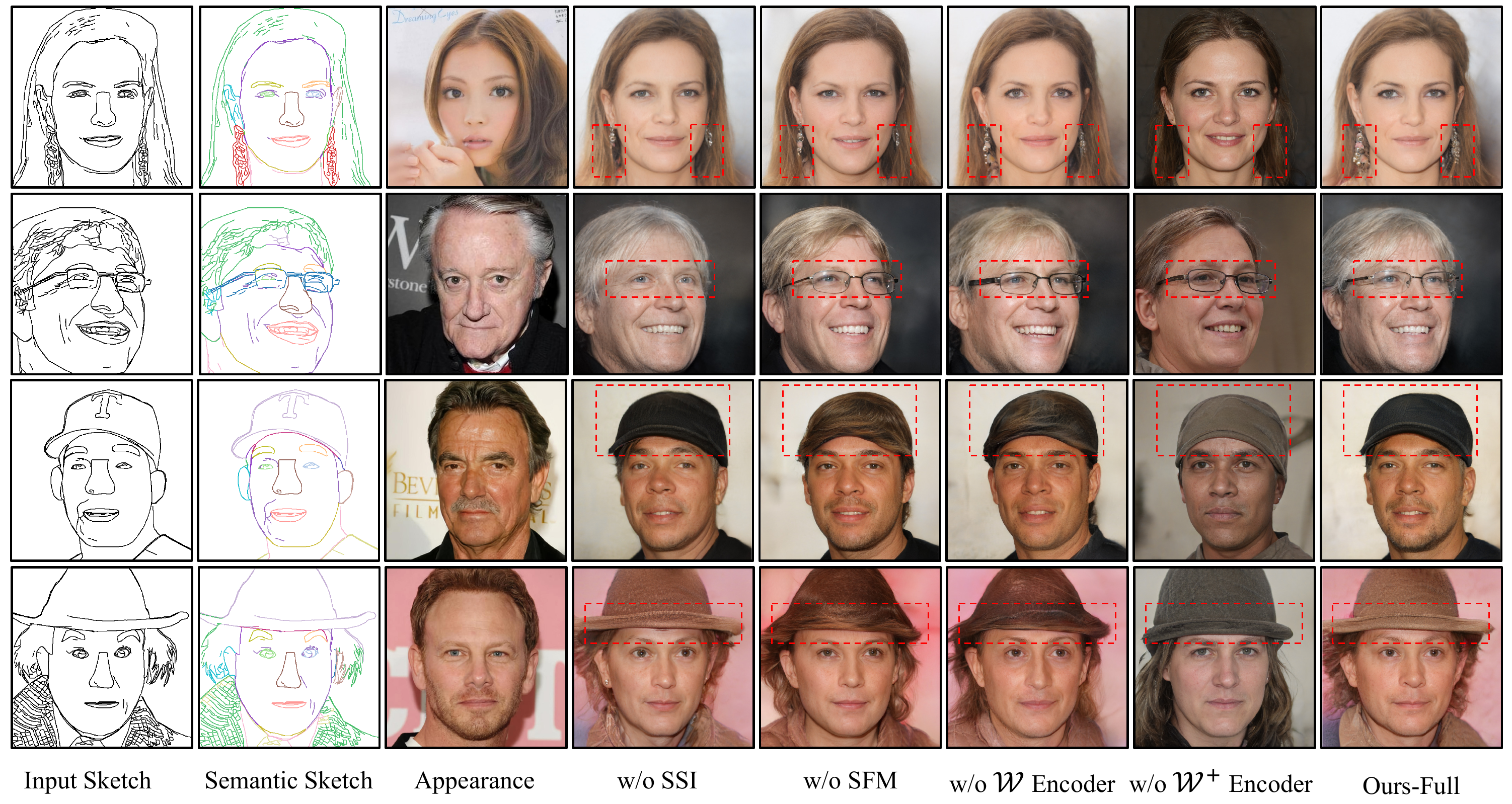}
\caption{Results of ablation study. To find out the impact of each component, we compare the results from 4 model variants and our full setting model. These 4 variants are designed by removing each key component from default setting.
}
\label{fig:ablation}
\vspace{-0.3cm}
\end{figure*}

Our framework consists of three key components, including the sketch semantic Interpretation, the $\mathcal{W}$-$\mathcal{W^+}$ encoder architecture, and the semantic feature matching loss.
We conducted ablation studies to demonstrate the impact of each of these three components.
\paragraph{Sketch Semantic Interpretation}
This module extracts stroke semantics from sketches.
We introduce a model variant, namely “w/o SSI”, by only changing the input of the second stage from a predicted semantic sketch image to a binary sketch image with the default settings.
By comparing the results from “w/o SSI” and “Ours-Full” in the first two rows of Fig.~\ref{fig:ablation}, it can be seen that the SSI module effectively captures all the sketch semantics, such as earrings and glasses.
This improvement is mainly attributed to the fact that the extracted semantic sketches are more identifiable than binary sketches.
Moreover, the SSI module narrows the gap caused by different levels of detail.
Specifically, as shown in Fig.~\ref{fig:ablation-SSI}, the “w/o SSI” model can generate earrings from binary sketches from the CelebA-Edge dataset while failing on those from the CelebA-Contour dataset with fewer details.
In contrast, our approach can generate more consistent results on both types of sketches, showing its generalization ability to sketches with different levels of detail.

\begin{table}[h]
\centering
\caption{Quantitative results on CelebA-Edge dataset for ablation study. We evaluate the image quality and semantic accuracy of synthesized results with P\_Acc, FID, and LPIPS.}
\begin{tabular}{@{}lccc@{}}
\toprule
Method                        & P\_Acc ($\uparrow$)        & FID ($\downarrow$)  & LPIPS ($\downarrow$)        \\ \midrule
w/o SSI         & 0.900  & 39.25   &  \textbf{0.430}    \\
w/o SFM         & \textbf{0.902}    & 44.33    & 0.463          \\
w/o $\mathcal{W}$ Encoder         & \textbf{0.902}    & 39.54  & 0.447           \\
w/o $\mathcal{W^+}$ Encoder    & 0.852    & 58.61    & 0.580          \\
Ours-Full              & \textbf{0.902}    &  \textbf{38.89}   & 0.446              \\ \bottomrule
\end{tabular}
\label{tab:quantitative_ablation_edge}
\end{table}

\begin{table}[h]
\centering
\caption{Quantitative results on CelebA-Contour dataset for ablation study. We evaluate the image quality and semantic accuracy of synthesized results with P\_Acc, FID, and LPIPS.}
\begin{tabular}{@{}lccc@{}}
\toprule
Method                        & P\_Acc ($\uparrow$)        & FID ($\downarrow$)  & LPIPS ($\downarrow$)        \\ \midrule
w/o SSI         & 0.891  & 44.10   &  \textbf{0.452}    \\
w/o SFM         & 0.879    & 47.94    & 0.480          \\
w/o $\mathcal{W}$ Encoder         & \textbf{0.892}    & 45.07  & 0.466           \\
w/o $\mathcal{W^+}$ Encoder    & 0.823    & 63.20    & 0.652          \\
Ours-Full              & \textbf{0.892}    &  \textbf{43.18}   & 0.466              \\ \bottomrule
\end{tabular}
\label{tab:quantitative_ablation_contour}
\vspace{-0.2cm}
\end{table}

\paragraph{The $\mathcal{W}$-$\mathcal{W^+}$ Encoder Architecture}
To evaluate our $\mathcal{W}$-$\mathcal{W^+}$ encoder on balancing the semantic controllability of $\mathcal{W}$ space and expressive power of $\mathcal{W^+}$ space, we introduce two model variants, namely ``w/o $\mathcal{W}$ Encoder" and ``w/o $\mathcal{W^+}$ Encoder", by removing $\mathcal{W}$ encoder and $\mathcal{W^+}$ encoders from our full model.
As shown in Table~\ref{tab:quantitative_ablation_edge} and \ref{tab:quantitative_ablation_contour}, the expressive power of $\mathcal{W^+}$ space makes ``w/o $\mathcal{W}$ Encoder" superior to ``w/o $\mathcal{W^+}$ Encoder" in terms of P\_Acc, LPIPS, and FID.
However, these quantitative metrics focus on the patchwise comparison of local shapes and textures. They do not always coincide with high-level semantic understanding. Without the $\mathcal{W}$ encoder, while the sketch is only embedded into the first 8 codes that represent coarse semantics, the model ``w/o $\mathcal{W}$ Encoder" frequently fails to associate correct fine textures with semantics. As the bottom two rows in Fig.~\ref{fig:ablation} show, even though the input sketches are mapped to the 'hat' in coarse semantics, the generator tends to add hair-like textures to the hat region.
On the other hand, when the input sketches are only mapped to the $\mathcal{W}$ space, the fine textures and coarse semantics are more coherent in the generated images. However, without the high DoF $\wplus$ encoder, the model ``w/o $\wplus$ Encoder" fails to capture all the semantics of minority parts, like earrings in the first example in Fig.~\ref{fig:ablation}. Moreover, this model sacrifices the fine-scale geometric alignment due to the low DoF $\mathcal{W}$ encoder. 
By combining the $\mathcal{W}$ encoder and $\wplus$ encoder to leverage their strengths on fine appearance coherence and semantics preservation, ``Ours-Full" achieves the best quantitative performance on P\_Acc and FID, as well as the visual quality. 
We further explore different combinations of the 18 codes to embed coarse-scale semantics and fine-scale appearance in the $\wplus$ space on the CelebA-Edge and CelebA-Contour datasets.
To better disentangle the semantics and textures in the image synthesis, we combine the semantics codes mapped from the input sketch and a randomly sampled appearance code in the latent $\wplus$ space to synthesize a face image. We compare the synthesized images with the corresponding ground-truth images and report the P\_Acc and FID in Table \ref{tab:quantitative_combination}.
As the number of codes used to embed coarse semantics for the sketches increases from 6 to 10, the semantics preservation performance indicated by P\_Acc increases accordingly. However, when more codes are used for sketch embedding and less codes for appearance, the performance drops due to the ambiguity of fine-level details in sketches. Especially for the CelebA-Contour dataset, the P\_Acc drops more from 10-8 combination to  the 12-6 combination than that on the CelebA-Edge dataset, while the contour-style sketches carry less fine-level information than edge-style sketches.
Considering FID which measures the image quality, the 8-10 combination, which is well-established for the pretrained generator, achieves the best results on both the CelebA-Edge and CelebA-Contour datasets. 
\begin{table}[h]
\centering
\caption{Results of different combinations of the 18 codes to encode sketch and appearance features in the $\wplus$ space. The first number in the first column denotes the number of codes used to embed coarse semantic features from sketches and the second number denotes the style codes used for fine appearance features.}
\begin{tabular}{@{}clcccc@{}}
\toprule
\multicolumn{2}{l}{\multirow{2}{*}{Method}} & \multicolumn{2}{c}{CelebA-Edge} & \multicolumn{2}{c}{CelebA-Contour} \\ \cmidrule(l){3-6} 
\multicolumn{2}{l}{}                         & P\_Acc ($\uparrow$)        & FID ($\downarrow$)  & P\_Acc ($\uparrow$)        & FID ($\downarrow$)           \\ \midrule
\multicolumn{2}{c}{6-12}   & 0.867    & 47.06       & 0.856            &   49.17            \\
\multicolumn{2}{c}{8-10}    &  0.893     &  \textbf{42.35}        & 0.883      &  \textbf{46.25}             \\
\multicolumn{2}{c}{10-8}  & \textbf{0.899} &   54.07    &   \textbf{0.888}          &  57.76             \\
\multicolumn{2}{c}{12-6}   & 0.898  &  52.41   &  0.884     &  61.32          \\\bottomrule
\end{tabular}
\label{tab:quantitative_combination}
\end{table}
\paragraph{The Semantic Feature Matching Loss}
The SFM loss is designed to provide supervision for the network to better capture the semantics of input sketches. 
As shown in Fig.~\ref{fig:ablation}, without $L_{SFM}$, the network fails to generate face images with accessories, such as hats, earrings, and glasses.
When quantitatively comparing the three metrics of ``w/o SFM" and ``Ours-Full" in Table~\ref{tab:quantitative_ablation_edge} and ~\ref{tab:quantitative_ablation_contour}, our full model performs better than ``w/o SFM" on both the CelebA-Edge and CelebA-Contour datasets. 
When comparing their performance on the two datasets, both models decline on the CelebA-Contour dataset due to the domain gap while the models are only trained with synthesized sketches in the style of CelebA-Edge dataset.  However, ``Ours-Full" model declines slightly from CelebA-Edge to CelebA-Contour, showing its generalization capability on different sketch styles by learning high-level semantics with the supervision of $L_{SFM}$.

\begin{figure*}[h]
\centering
\includegraphics[width=0.95\textwidth]{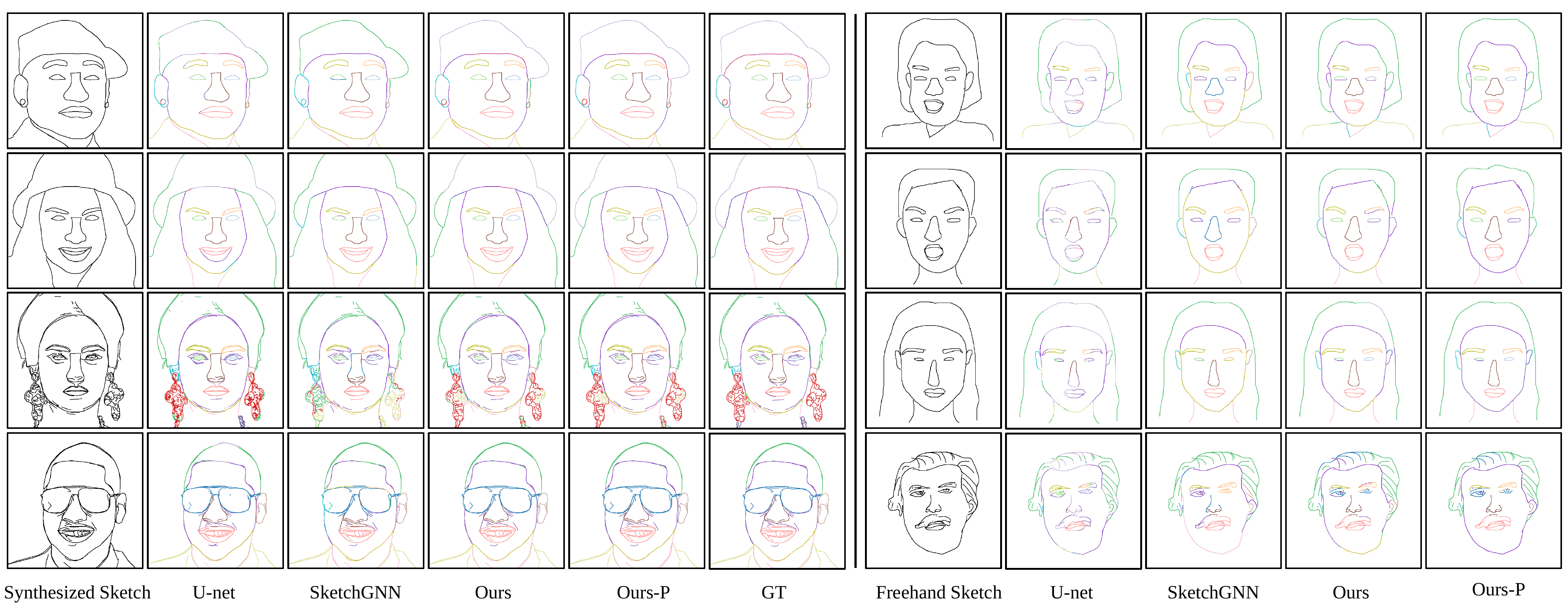}
\caption{Qualitative comparison of semantic prediction results on synthetic sketches from the CelebA-Edge and CelebA-Contour datasets(left) and freehand sketches (right). Compared to SketchGNN~\cite{lumin2021sketchgnn} and U-Net~\cite{unet}, our SSI module produces comparable results on synthetic sketches, while outperforming them on freehand sketches that are much coarser than synthetic sketches. Our post-processing step further enforces the semantics consistency in long strokes (Ours-P).}
 
\label{fig:SSI-precision}
\vspace{-0.3cm}
\end{figure*}

\subsection{Perceptual Study}
In each test, the input sketch is shown in the middle, while the synthesized results by our method and one of the compared methods are displayed on two sides in random order. 
The subjects were asked to select a favored result considering which is more realistic or more similar to the input sketch.
$25$ subjects participated in our user study and each subject worked on $50$ image triplets. Table \ref{tab:user_study} gives the results.  
In regard to the image quality, our method is considered significantly better than pix2pixHD and DeepFaceDrawing on both datasets. Compared with pSp, our method does not show significant superiority since they both use the pretrained generator to synthesize face images. 
In regard to the fidelity to sketches, our method works slightly better than pix2pixHD on CelebA-Edge and significantly better on CelebA-Contour. This is mainly because pix2pixHD requires the sketch style similar to the training sketches and fails to generalize on contour-style sketches. 
Compared with DeepFaceDrawing, our method is superior in preserving the sketch semantics and structures since DeepFaceDrawing loses fine control during manifold projection.
We do not compare with DeepFaceEditing because it works surprisingly poorly on our sketch data. We deduce the reason is that it emphasizes patch-wise correspondence between the generated images and input sketches for local editing, thus failing to generalize to freehand sketches that exhibit large geometric distortions. 
Our method narrowly wins the competition with pSp and ReStyle on CelebA-Edge dataset since most test cases are common faces without accessories presented. 
But the considerable superiority of our method on both datasets still demonstrates the effectiveness of our sketch semantics embedding.
It is worth noting that our approach has a greater advantage on the contour dataset in terms of the fidelity to input sketches. This is mainly because conveying precise semantics is more critical for contour-style sketches that are more sparse than edge maps that have rich details.

\begin{table}[h]
\caption{Perceptual study results of visual quality and semantics fidelity to sketches on the CelebA-Edge and CelebA-Contour datasets. The numbers denote the percentage of participants who favored our method over the other one.}
\centering
\begin{tabular}{@{}lcccc@{}}
\toprule
\multirow{2}{*}{Method} & \multicolumn{2}{c}{Quality} & \multicolumn{2}{c}{Similarity} \\ \cmidrule(l){2-5} 
 & Edge & Contour & Edge & Contour \\ \midrule
Ours vs. pix2pixHD  & $80.8\%$ & $100.0\%$  &  $61.0\%$ & $94.5\%$    \\ 
Ours vs. DeepFaceDrawing  & $95.7\%$ & $89.7\%$ & $89.2\%$ & $86.6\%$   \\
Ours vs. pSp & $57.5\%$ & $56.7\%$  & $54.4\%$ & $60.3\%$ \\ 
Ours vs. ReStyle & $52.5\%$ & $97.6\%$  & $53.3\%$ & $95.3\%$ \\ 
\bottomrule
\end{tabular}
\label{tab:user_study}
\vspace{-0.3cm}
\end{table}

\subsection{Precision of Sketch Semantic Interpretation}
To further validate the performance of our SSI module, we conduct comparison experiments with SketchGNN\cite{lumin2021sketchgnn} and U-Net~\cite{unet}.
All comparing methods are trained only on the CelebA-Edge dataset.
As shown in Table~\ref{tab:SSI_Accuracy}, our SSI module outperforms SketchGNN on both CelebA-Edge and CelebA-Contour datasets.
The performances of U-Net and our SSI module are comparable on the CelebA-Edge dataset while U-Net achieves slightly higher accuracy than our SSI module on the CelebA-Contour dataset.
\begin{table}[h]
\caption{Accuracy of semantic prediction.}
\centering
\begin{tabular}{lccc}
\toprule
Method  & SSI  & U-net & SketchGNN \\
\midrule
Edge  & $91.2\%$  &  $91.3\%$  &  $86.0\%$  \\ 
Contour  & $90.9\%$ & $96.6\%$  & $87.0\%$ \\
\bottomrule
\end{tabular}
\label{tab:SSI_Accuracy}
\vspace{-0.2cm}
\end{table}

\begin{figure}[h]
\centering
\includegraphics[width=0.9\columnwidth]{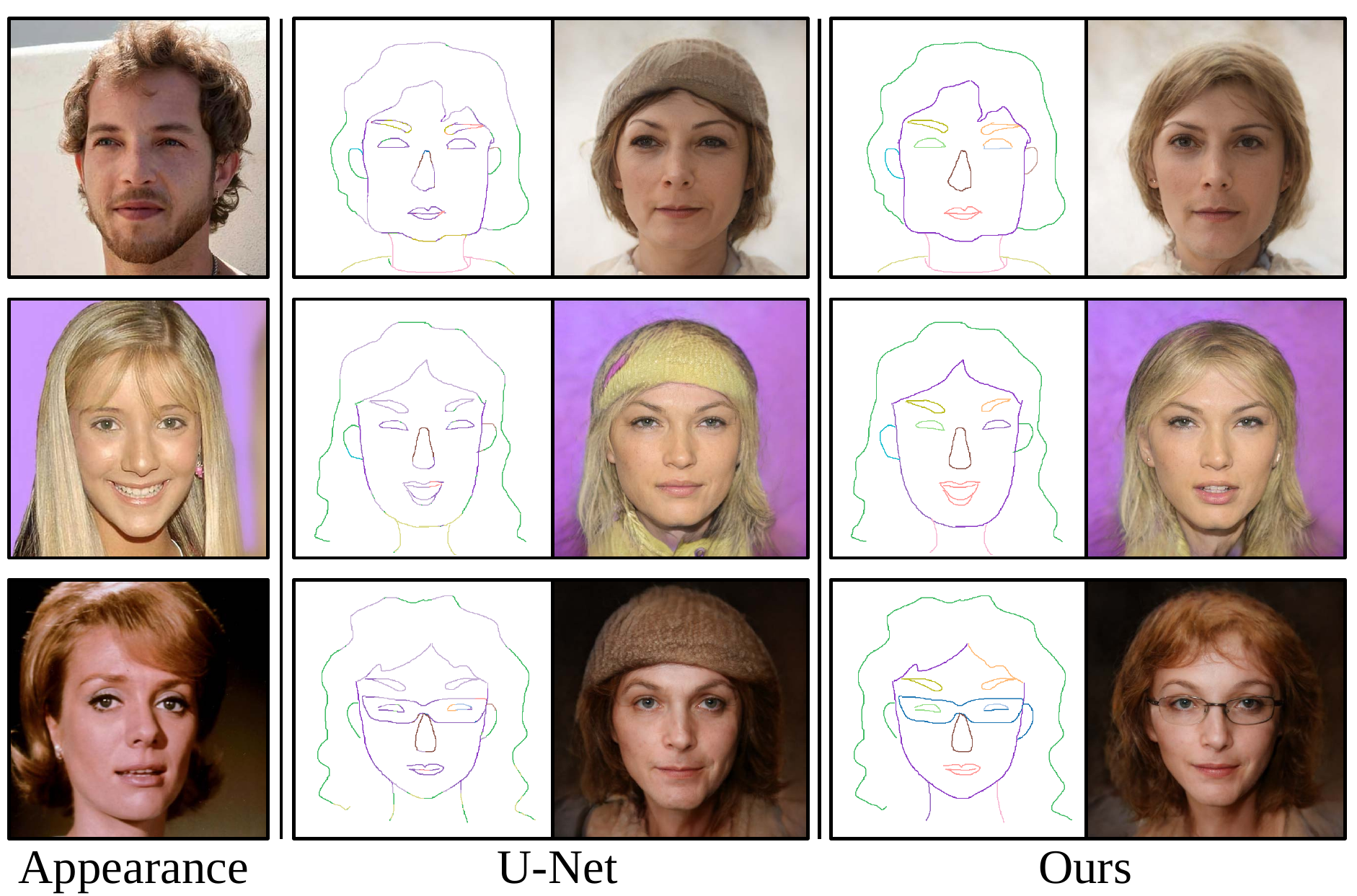}
\caption{Visual results of the predicted semantic freehand sketches and corresponding generated face images. }
\label{fig:ablation-unet}
\vspace{-0.3cm}
\end{figure}

But for freehand sketches, SketchGNN and U-Net both fail to predict correct semantics as shown in Fig.~\ref{fig:SSI-precision}.
Since our SSI module utilizes stroke structure and topology relationships between strokes for semantic prediction, our method exhibits better generalization ability on freehand sketches.
Particularly, each long stroke drawn by a user usually belongs to only one semantic category but it is cut into a set of short segments for semantics interpretation. Therefore, we assign all the short strokes in the same long stroke with the same semantic label via a voting strategy. The user can optionally choose to use the semantic sketches with post-processing or not.
In Fig.~\ref{fig:SSI-precision}, we also compare the semantic sketches before and after post-processing to demonstrate that this post-processing can generate more consistent semantic interpretation for long strokes.
Furthermore, we feed the predicted results from U-Net and the predicted results with post-processing from our SSI module into our sketch embedding module for face generation to investigate how semantic prediction performance affects the final face generation.
Given sketches that share the same geometry but with different semantics, our $\mathcal{W}-\mathcal{W^+}$ encoder can accurately convey the semantics to the synthesized faces, as shown in Fig.~\ref{fig:ablation-unet}. 
While U-Net predicts semantic labels for the hair as hat and the glasses as skin, the synthesized faces from our sketch semantic embedding complies the given sketch semantics.

\begin{figure*}[t]
\centering
\includegraphics[width=0.9\textwidth]{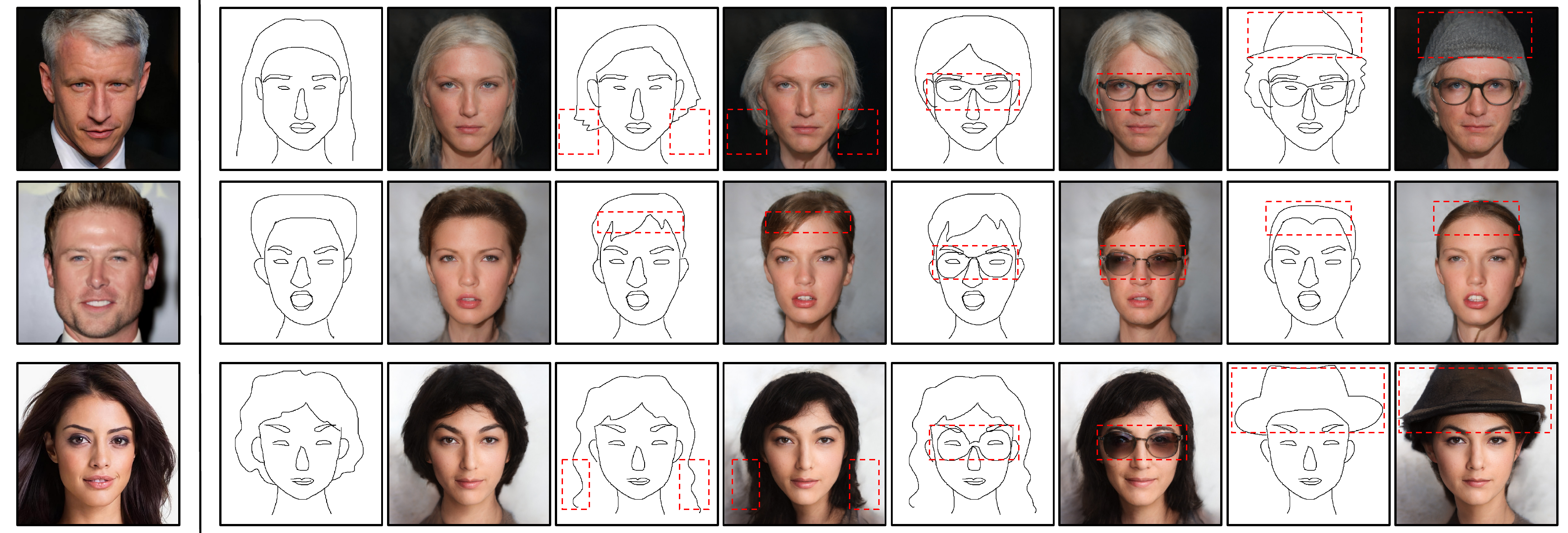}
\caption{Results of local face editing. 
Despite the high abstract level of freehand sketches, our approach still supports precise local editing.
Our method controls the geometric and semantic information of the generated face images through sketches without the interference of appearance.
}
\label{fig:editing}
\vspace{-0.3cm}
\end{figure*}

\begin{figure}[t]
\centering
\includegraphics[width=0.9\columnwidth]{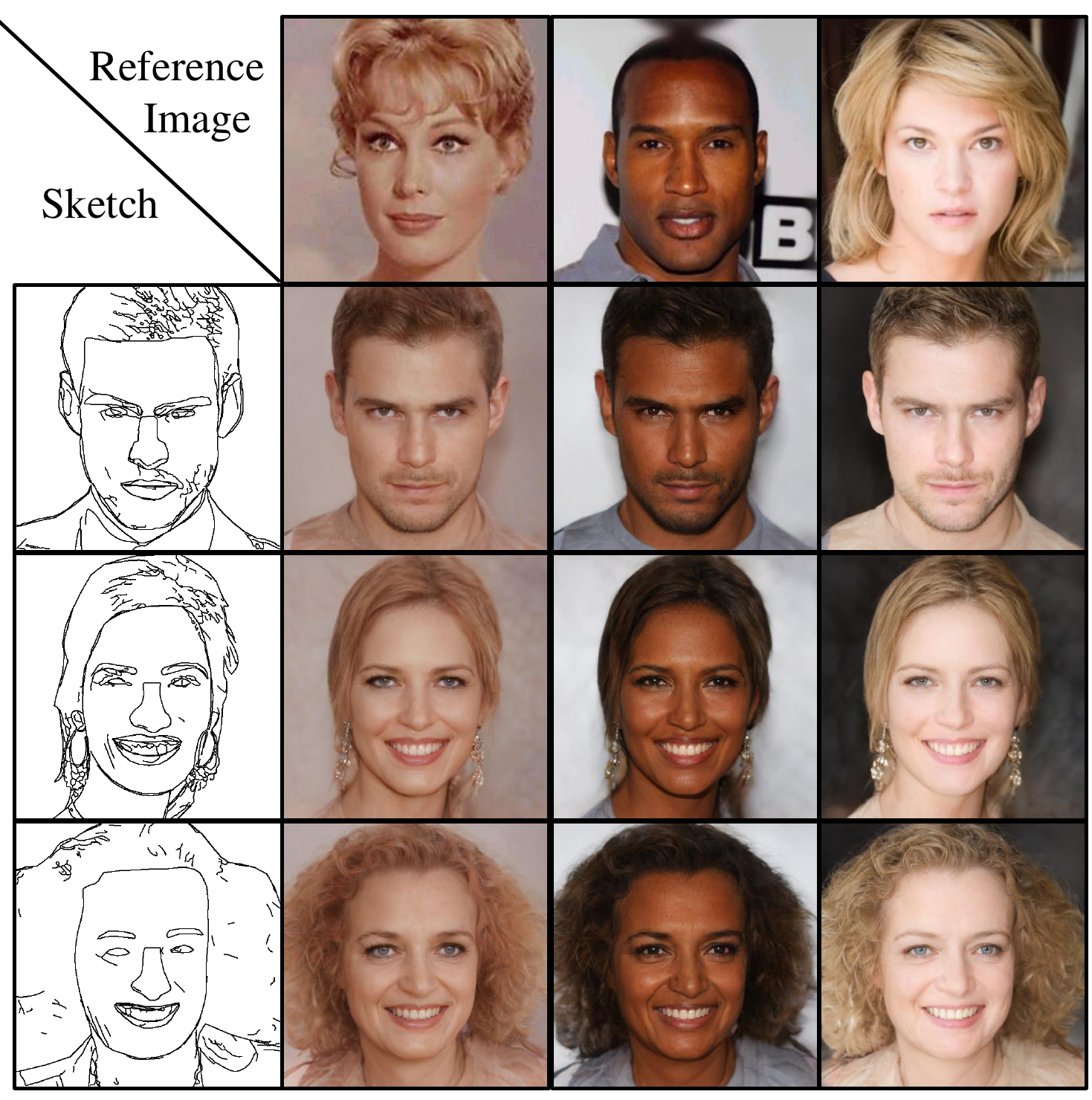}
\caption{Results of appearance manipulation.
Our method effectively disentangles the influence of sketches and reference images on the synthesized face images.}
\label{fig:appearance_manipulation}
\vspace{-0.3cm}
\end{figure}
\begin{figure*}[h]
\centering
\includegraphics[width=0.9\textwidth]{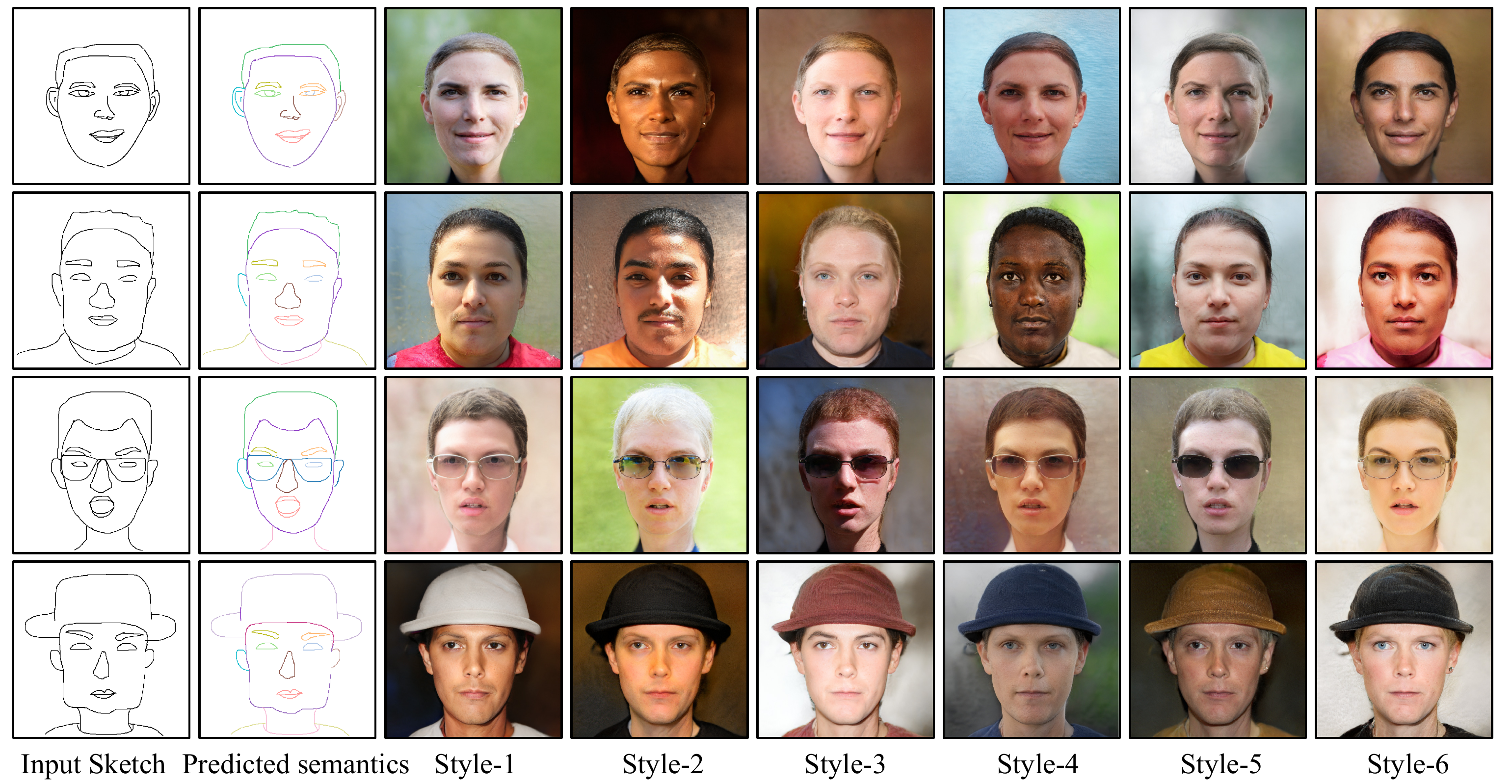}
\caption{Sketch-based face generation with random appearance codes. Given the same sketch, our method can generate face images with diverse appearances from noise vectors.} 
\label{fig:appearance_diversity}
\vspace{-0.2cm}
\end{figure*}
\begin{figure*}[t]
\centering
\includegraphics[width=0.9\textwidth]{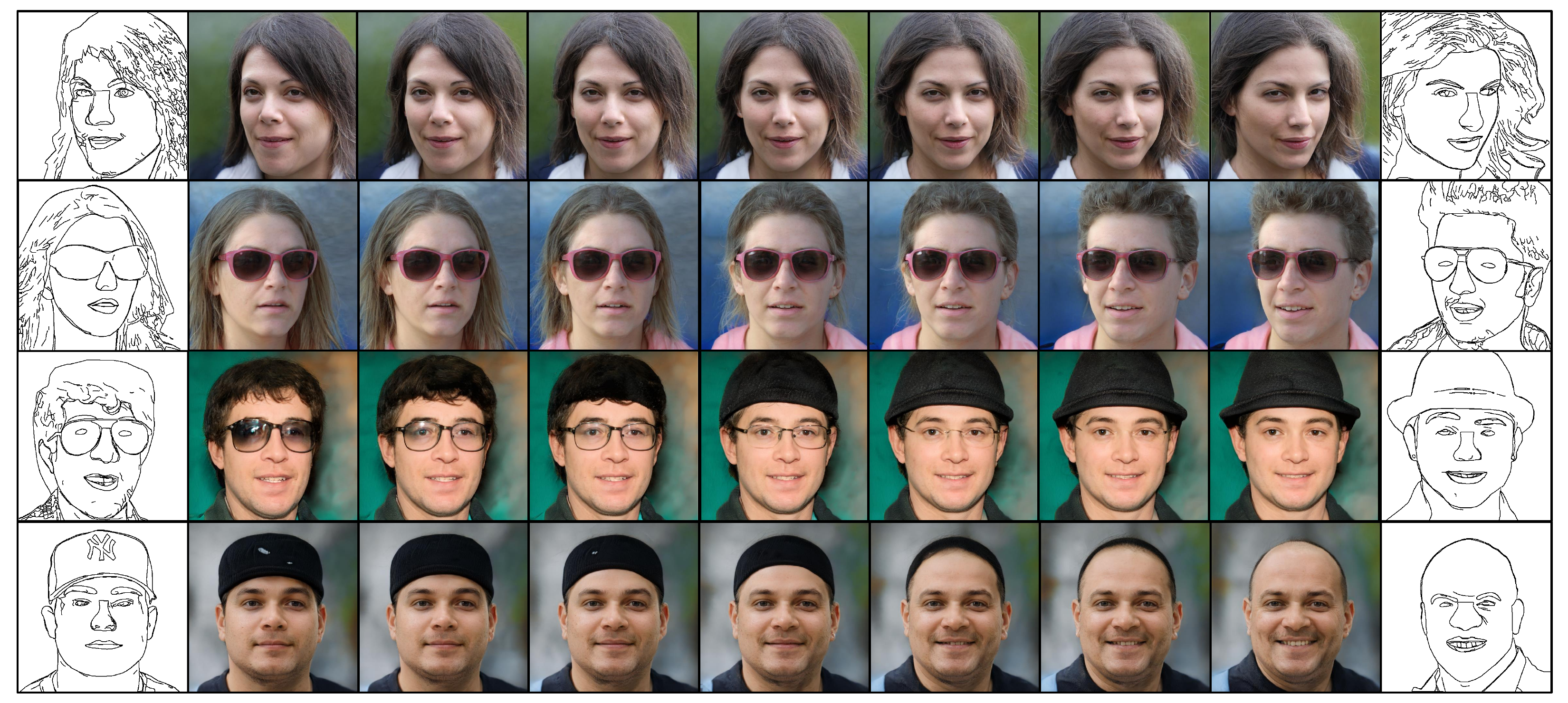}
\caption{Face interpolation. Results of face interpolation. Given two sketches, our method can smoothly generate intermediate face images by interpolating the two latent vectors encoded from two sketches and feeding them to a face generator. 
} 
\label{fig:morphing}
\vspace{-0.3cm}
\end{figure*}


\section{More Applications}
With a TitanXP GPU and an i7-6800k CPU, our method averagely takes 24.91ms to predict stroke semantics and 218.32ms to generate a 1024×1024 face image, thus well supporting interactive face synthesis.
Besides directly generating face images with a sketch, our method is flexible and powerful to support a wide range of applications, including local editing, appearance manipulation, and face morphing. We design a web user interface that supports these applications, as shown in the supplementary video. 

\paragraph{Local Face Editing}
Our framework supports local face editing with fine controls by sketches, including adding decorative accessories, changing hairstyles, changing hats, etc.
Even for large shape modifications, our framework still achieves precise local editing without affecting the other parts, as shown in Fig.~\ref{fig:editing}.

\paragraph{Appearance Manipulation}
Given an input sketch, generating face images with consistent geometry and semantics with arbitrary appearances or specified appearances from a reference image is beneficial for many applications.
Fig.~\ref{fig:appearance_manipulation} shows the results of using different reference images to control the appearance of generated faces given the same sketch.
Our method effectively disentangles the influence of the sketches and the reference appearances on the generated images.
Our framework also supports generating diverse appearances for the same sketch by mixing the sketch code with a random appearance code, as Fig.~\ref{fig:appearance_diversity} shows.

\paragraph{Face Interpolation}
Given two sketches, we can obtain their corresponding latent vectors $\mathbf{w}_1,\mathbf{w}_2$ from our sketch encoder $E_S$. 
By interpolating these two vectors and then feeding the intermediate vectors to the pretrained face generator, we can do face interpolation.
From the results shown in Fig.~\ref{fig:morphing}, one can see that our method can smoothly change the high-level semantics, including head pose, gender, face shape, and even decorative accessories.

\section{Conclusion}
In this work, we present a novel sketch-based semantics-preserving face generation framework that consists of a semantics-preserving sketch embedding module and a sketch semantic interpretation module.
The proposed \wwe encoder architecture balance the semantic controllability and perceptual quality of $\mathcal{W}$ space and reconstruction quality of $\mathcal{W^+}$ space, and shows superiority in precisely mapping the stroke semantics in the latent space for high-quality face synthesis.
Moreover, a semantic feature matching loss is designed to provide effective supervision on semantics embedding.
To alleviate the difficulty of semantics-preserving sketch embedding module in implicitly capturing semantics from binary sketches, a sketch semantic interpretation module is designed to use stroke structure and topological relationship between strokes for semantic prediction.
The sketch semantic interpretation module presents great generalization ability to various types of sketches.
Extensive experiments conducted with sketches in various styles prove the superiority of our method in generating semantically consistent results with the input sketches.
Moreover, our method also shows better generalization ability to freehand sketches and disentanglement ability between geometry, semantics, and appearance.

Since the latent space is low-dimensional and highly compact, fine-scale geometric information from sketches is inevitably lost. As a result, the outputs may not be geometrically strictly aligned with the input sketches.
Though this mapping is not conformal in local geometry, it is beneficial for non-expert users who draw sketches with varying degrees of abstraction. If the results are exactly geometrically consistent with the coarse sketches that have geometric distortions, the synthesized results would be geometrically incorrect and unrealistic.
Moreover, our approach controls the global appearance of outputs by a reference face image. Therefore, our model cannot achieve local appearance manipulation. For example, the color of the hat and hair cannot be separably controlled.

\bibliographystyle{IEEEtran}
\bibliography{sample-bib}

\begin{IEEEbiography}
[{\includegraphics[width=1in,clip,keepaspectratio]{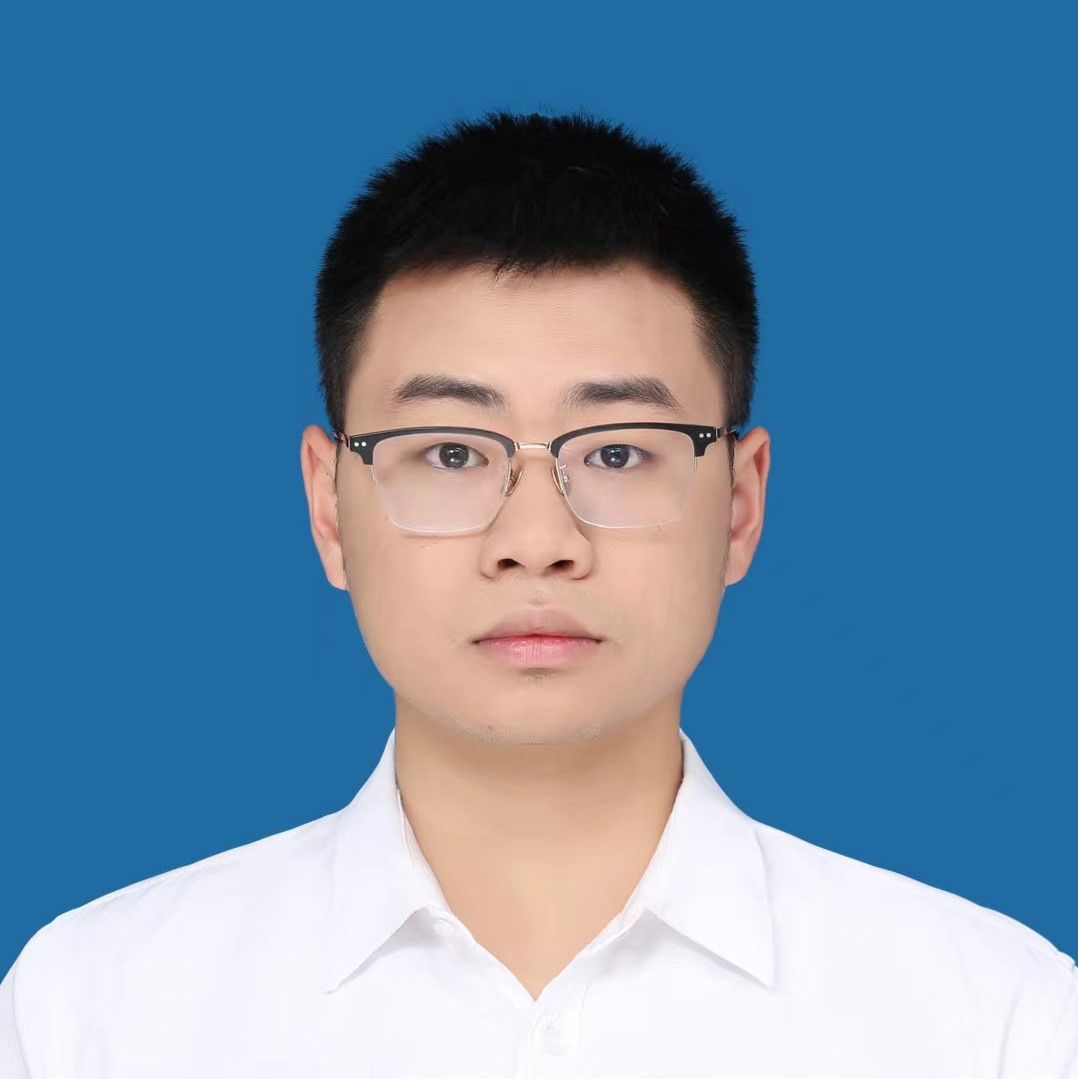}}]
{Binxin Yang} received the B.S. degree in electronic science and technology from University of Science and Technology of China in 2019, where he is
currently pursuing the Ph.D. degree with the School of Information Science and Technology, being advised
by Xiaoyan Sun and Xuejin Chen. 
His research interest involves 2D image synthesis, especially face generation.
\end{IEEEbiography}
\begin{IEEEbiography}
[{\includegraphics[width=1in,clip,keepaspectratio]{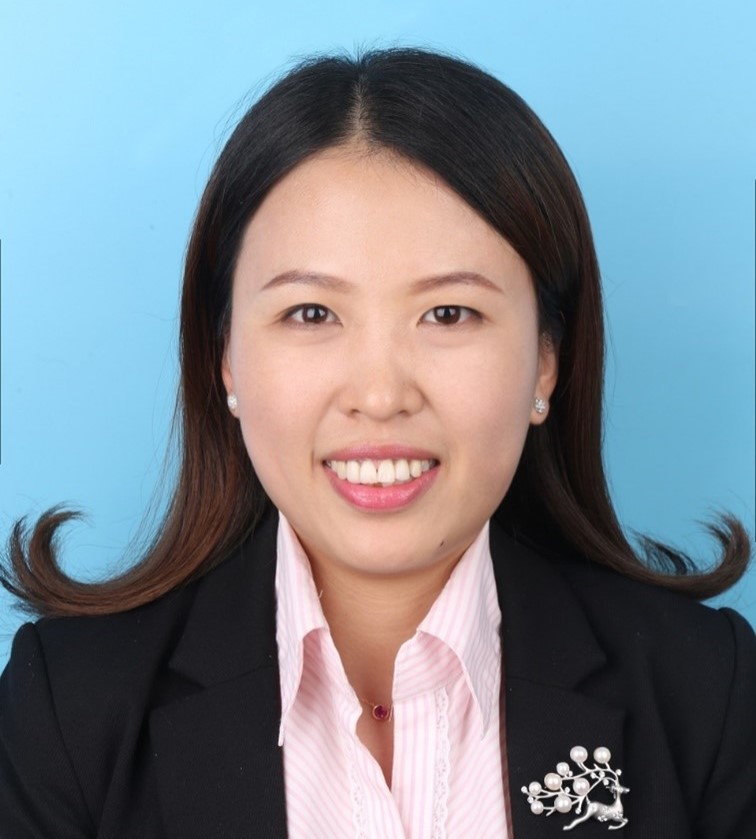}}]
{Xuejin Chen} received the B.S. and Ph.D. degrees in electronic circuits and systems from the University of Science and Technology of China, Hefei, China, in 2003 and 2008, respectively. From 2008 to 2010, she conducted research as a Post-Doctoral Scholar with the Department of Computer Science, Yale University, New Haven, CT, USA. She is currently a Professor with the School of Information Science and Technology, University of Science and Technology of China. Her research interests include 3-D modeling, geometry processing, and content creation. She has authored or co-authored over 70 articles in these areas. Dr. Chen was one of the recipients of the Honorable Mention Awards of Computational Visual Media in 2019.
\end{IEEEbiography}

\begin{IEEEbiography}
[{\includegraphics[width=1in,clip,keepaspectratio]{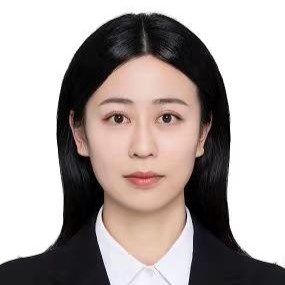}}]
{Chaoqun Wang} received the B.S. degree in department of electronic engineering and information science from University of Science and Technology of China, in 2018. She is currently studying for Ph.D. in USTC. Her major research
interests span zero-shot learning, cross-modal learning, and inpainting.
\end{IEEEbiography}
\begin{IEEEbiography}
[{\includegraphics[width=1in,clip,keepaspectratio]{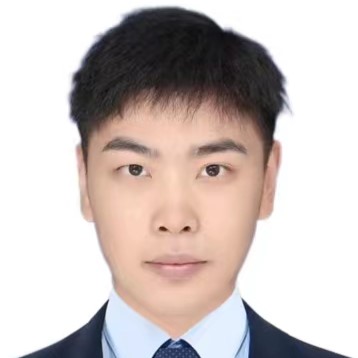}}]
{Chi Zhang} received his MSc of computer science from University of Science and Technology of China. He is currently a researcher with ByteDance Inc. His research focuses on multimodal content understanding, 3D reconstruction, and medical image understanding.
\end{IEEEbiography}
\begin{IEEEbiography}
[{\includegraphics[width=1in,clip,keepaspectratio]{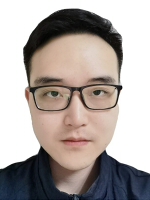}}]
{Zihan Chen} received the B.S. degree in electronic information engineering from Guilin University of Technology of China, in 2019. He is currently pursuing the Ph.D. degree in information and communication engineering from the University of Science and Technology of China, Hefei.
His research interest involves 2D image synthesis, especially image-to-image translation.
\end{IEEEbiography}
\begin{IEEEbiography}
[{\includegraphics[width=1in,clip,keepaspectratio]{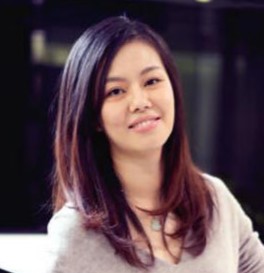}}]
{Xiaoyan Sun} is a Full Professor at University of Science and Technology of China and 
deputy director of National Engineering Laboratory for Brain-inspired Intelligence Technology and 
Application. Prior to joining USTC in 2019, she was with Microsoft Research Asia from 2003 to 2019, 
most recently as a Senior Researcher. Xiaoyan Sun received the B.S., M.S., and Ph.D. degrees in 
computer science from Harbin Institute of Technology, Harbin, China, in 1997, 1999, and 2003, 
respectively. She had been an intern in Microsoft Research Asia since 2000 before joining Microsoft 
Research Asia in 2003. Her research interests include computer vision, image/video processing, 
machine learning and artificial intelligence. She has authored or co-authored over 100 publications in 
journals and conferences, 10 proposals to standards with one accepted, and hosts over 10 granted US 
patients. She was a recipient of the Best Paper Award of the IEEE Transactions on Circuits and Systems 
for Video Technology in 2009 and the Best Student Paper Award of VCIP 2016. She is on the Senior 
Editorial Board of the IEEE Journal on Emerging and Selected Topics in Circuits and Systems and AE of 
Signal Processing: Image Communication.
\end{IEEEbiography}

\end{document}